\documentclass[letterpaper,10pt,journal,twoside]{IEEEtran}

\IEEEoverridecommandlockouts

\usepackage{cite}
\usepackage{amsmath,amssymb,amsfonts}
\usepackage{graphicx}
\graphicspath{{figures/}}
\usepackage{booktabs}
\usepackage{tabularx}
\usepackage{multirow}
\usepackage{float}
\usepackage[linesnumbered,ruled,lined]{algorithm2e}

\usepackage{amsthm}
\usepackage{tikz}
\usetikzlibrary{positioning}
\usepackage{hyperref}

\providecommand{\Description}[1]{}

\title{Model-Based Imitation Learning For Monocular Vision-Based Navigation}
\author{Basant Sharma$^{1}$, Prajyot Jadhav$^{1}$, Saqi Hussain Kalan$^{1}$, Arun Kumar Singh$^{1}$%
\thanks{This work was in part supported by grant PSG753 from Estonian Research Council,
project TEM-TA101 funded by European Union and Estonian Research Council and SekMO program (2021-2027.1.01.23-0419) co-funded by European Union.}
\thanks{$^{1}$:Institute of Technology, University of
Tartu, Tartu, Estonia.}
}

\begin{document}
\maketitle

\begin{abstract}
Reliable navigation with a single monocular limited field-of-view camera is challenging
because the lack of depth measurements and limited spatial awareness complicate
collision checking. Existing approaches to tackle these challenges fall into two
categories. Model-based planners can incorporate explicit collision-avoidance guidance
which leads to better generalization but are often computationally expensive and
require manual tuning of the algorithmic parameters. In contrast, end-to-end approaches
trained by imitation learning can be flexible but typically encode safety only
implicitly and thus can suffer from high collision rates in novel scenarios.

In this paper, we present a middle ground through model-based imitation learning. The
model-based part is a learned collision model that can map a sequence of past images and
a candidate control sequence to a distribution of worst-case obstacle clearance. This
predictive distribution in turn is used to define a collision-risk cost in a trajectory
optimization formulation. The imitation component uses human demonstrations to learn
scene-dependent balancing of collision risk with other task costs such as goal progress.
We also use it to learn the algorithmic parameters (e.g., a warm start) of our optimizer to
accelerate the solution process. We show that our approach can demonstrate highly
reliable navigation in cluttered environments on robots with different embodiments. We
also demonstrate a substantial reduction in collision rate over recent end-to-end and
model-based baselines.
\end{abstract}


\section{Introduction}
\label{sec:introduction}
Reliable autonomous navigation capability in highly cluttered environments is a
fundamental requirement for deploying mobile robots and requires 3D perception of the
environment. For some robot embodiments such as quadrotors, using the typical 3D sensors
such as LiDAR proves challenging from the perspective of payload capacity or energy
consumption. Thus, this work aims to recreate LiDAR-like navigation with just a single
monocular camera with a limited field of view. The core challenge in this regard stems from the
lack of metric depth information from native RGB observations. Moreover, because of
the limited FOV, the robot can only perceive a part of the environment at any given time and thus
can lead to additional collision risk.

Existing approaches to monocular visual navigation broadly fall into model-based and
end-to-end learning-based paradigms. The former class of methods relies on using estimated
depth~\cite{depthanything,NEURIPS2024_26cfdcd8,https://doi.org/10.48550/arxiv.2302.12288}
to build occupancy maps \cite{simon2023mononav} or subsequently plan using these
representations. This structure allows for explicit reasoning about collision avoidance
but monocular depth estimates can be noisy and the distribution of the noise can change
with viewing angle and lighting conditions. MonoMPC~\cite{11283048}
explicitly accounts for the depth estimation uncertainty and shows stronger performance
using a sampling-based MPC approach. However, its underlying planner is still
computationally demanding and requires extensive manual tuning. For example, the weights
of the cost function need to be manually tuned and the performance of the
sampling-based trajectory optimizer depends critically on algorithmic parameters such as
the initial distribution and the temperature parameter.

End-to-end approaches avoid constructing explicit occupancy maps or collision models.
Instead, they directly map sequences of images and goal context to short-horizon
trajectories or control inputs \cite{10610665}. FlowNav~\cite{gode2025flownav}
uses depth features from the current image as additional conditioning input
to improve reliability. These approaches are very flexible and can directly learn
navigation behavior through human demonstrations. However, collision
avoidance is generally encoded implicitly through demonstrations or introduced through
auxiliary safety corrections rather than incorporated directly into the
trajectory-generation objective. As a result, their performance can be brittle,
especially in environments which differ substantially from the training data. One simple
alternative used in works like \cite{kim2025care} is to augment end-to-end approaches
with a safety filter based on the estimated depth. But this cited work does not include
depth uncertainty and we observed that the performance gain provided by it can be
marginal(see supplementary material).

\noindent \emph{\textbf{Our Approach and Contributions}:} In this paper, we present a model-based
imitation-learning approach for monocular image-based navigation. Our approach can be
viewed as the middle ground between purely model-based planners and end-to-end
approaches. It inherits the safety assurances of the former while also being capable of
adapting the navigation behavior by learning from human demonstrations.

Our first contribution, which represents the \emph{model-based} aspect of our approach,
is a probabilistic collision model that maps sequences of monocular observations to
trajectory-dependent collision risk. Like FlowNav~\cite{gode2025flownav} and
MonoMPC~\cite{11283048}, we use Depth Anything V2~\cite{NEURIPS2024_26cfdcd8}.
Our model converts depth estimates from the image sequence into point-cloud
embeddings and concatenates them with the control sequence to obtain
a manipulatable distribution of worst-case clearance
and subsequently collision risk. We show that the encoded temporal context allows our
learned model to reason about collision with the obstacles currently not present in the
field of view. This provides our learned collision model with a sense of memory without
explicitly needing to build an occupancy map.

Our second contribution is a learned planning architecture that embeds a parameterized,
risk-aware trajectory optimizer directly within the neural policy. The resulting local
navigation policy consists of two coupled components: a neural parameter-prediction
network and a trajectory optimizer. The neural module maps visual observations and
start-goal context to two sets of parameters. The first set consists of the weights that
balance the trade-off between minimizing collision risk and other task-relevant
costs such as goal progress. The second set consists of algorithmic hyperparameters that
configure the trajectory optimizer itself and accelerate its convergence to a
locally-optimal solution.

To make this architecture trainable from demonstrations, we formulate the trajectory
optimizer as a differentiable soft Cross-Entropy Method (soft-CEM), motivated by the
differentiable CEM formulation of Amos and Yarats~\cite{pmlr-v119-amos20a}. The differentiability allows
the imitation loss to be mapped back to the weights of the neural module. The resulting
policy therefore learns an end-to-end mapping from visual observations and robot
states to the parameters that configure trajectory optimization, while preserving an
explicit model-based planning structure. Throughout the paper, we refer to the neural
parameter-prediction component as the \emph{policy network}.

\noindent\emph{\textbf{Practical Benefits and Experimental Results}:}
Our approach offers three key benefits: it provides explicit collision-risk awareness during trajectory generation, leverages observation history to overcome limited-FOV constraints for smoother navigation, and automates cost weight and hyperparameter tuning to reduce latency and manual effort.

We validate the above claimed benefits in real-world experiments against both end-to-end
and model-based baselines. For the former, we chose FlowNav~\cite{gode2025flownav}. In a Corridor-like environment, our method achieves an observed collision rate of $4\%$ at an average speed of $0.80$\,m/s, compared with collision
rate of $50\%$ for FlowNav. We use MonoMPC as the model-based
baseline. Across the three environments used for comparison with MonoMPC, our method
achieves collision rates of $0\%$--$7\%$, compared with $23\%$--$70\%$ for MonoMPC. Our approach also produces smoother motion, with an average excess
heading variation of $0.0188$\,rad/step, compared with $0.0306$\,rad/step
for MonoMPC.

\section{Related Works}
\noindent We present a brief description of how our work improves upon the state-of-the-art.

\subsection{Temporal Mapping for Visual Navigation}
The Cognitive Mapping and Planning (CMP) approach~\cite{gupta2017cognitive} learns a top-down spatial memory
for differentiable planning, while Active Neural SLAM~\cite{chaplot2020learning}
predicts obstacle maps and robot pose. Semantic MapNet~\cite{cartillier2021semantic}
aggregates RGB-D observations into semantic maps. For monocular navigation,
MonoNav~\cite{simon2023mononav} fuses estimated depth into a 3D reconstruction
and searches over motion primitives. These methods use temporal information
to construct spatial representations. \emph{Our approach instead predicts a
distribution of worst-case obstacle clearance for each candidate control
sequence. This allows previously observed obstacles outside the current view
to inform trajectory-level risk assessment without constructing an explicit
occupancy or semantic map.}

\subsection{Learning Collision Models}
Kahn et al.~\cite{kahn2018self} learn action-conditioned collision probabilities
from image sequences, supervised by collision outcomes recorded during
robot interaction. N-MPC~\cite{jacquet2024n} predicts collision scores at queried 3D positions from sensor depth images and uses these scores to constrain predicted positions in MPC. \emph{We differ from Kahn et al.~\cite{kahn2018self} because we learn clearance distributions instead of direct probabilities. Supervised clearance labels are easier to obtain and can also be generated for unexecuted candidate trajectories.} Our work is also closely related to MonoMPC~\cite{11283048}, which predicts a worst-case clearance distribution from the current image and uses it in risk-aware MPC. \emph{We extend this formulation with
image history and one-sided losses that directly supervise the risk surrogate
for safe and unsafe trajectories. As a result, our model is able to predict reliable collision risk even for obstacles which are not currently in the robot's camera view.}

\subsection{Model-Based Imitation Learning}
\label{sec:related_model_based_il}

Universal Planning Networks~\cite{pmlr-v80-srinivas18b} and
Pan et al.~\cite{pan2020zero} learn latent predictive models through
imitation losses on optimized actions.
MILE~\cite{hu2022mile} jointly learns a visual world model and a
driving policy.
Differentiable MPC~\cite{amos2018differentiablempc} learns controller
costs and dynamics from demonstrations, while
DriViDOC~\cite{acerbo2024driving} learns image-conditioned MPC cost
parameters.
DIPP~\cite{huang2024dipp} jointly learns agent predictions, cost
weights, and planner initialization through a differentiable planner,
using prediction and planning losses.

\emph{Whereas UPN~\cite{pmlr-v80-srinivas18b} and DIPP~\cite{huang2024dipp}
learn predictive models through differentiable planning, we train our collision
model separately and freeze it during policy learning. This prevents policy training from reducing the risk loss by modifying the collision model rather than improving the planned motion.}

\section{Methods}
\subsubsection*{Symbols and Notation}
\noindent We will use lowercase letters to represent scalars, while bold
font variants will be used for vectors. Matrices will be represented by uppercase bold
font letters.


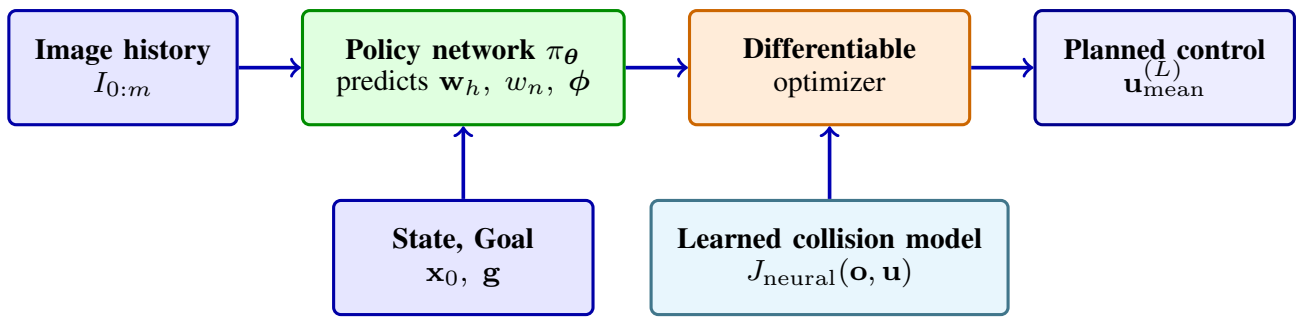
\begin{figure}[!htbp]
\centering
\resizebox{0.47\textwidth}{!}{%
\begin{tikzpicture}[
    font=\footnotesize,
    node distance=5.5mm and 6mm,
    base/.style={rounded corners=2pt,align=center,minimum height=11mm,
                 inner xsep=2.5mm,inner ysep=2.0mm,line width=0.85pt},
    input/.style={base,draw=blue!65!black,fill=blue!10},
    fixed/.style={base,draw=cyan!45!black,fill=cyan!8},
    policy/.style={base,draw=green!55!black,fill=green!12},
    optimizer/.style={base,draw=orange!80!black,fill=orange!15},
    output/.style={base,draw=blue!55!black,fill=blue!7},
    forward/.style={->,draw=blue!70!black,line width=1.0pt},
    backward/.style={->,draw=red!70!black,line width=1.0pt,densely dotted}
]

\node[input, minimum width=22mm] (images)
    {\textbf{Image history}\\ $I_{0:m}$};

\node[policy, right=of images, minimum width=31mm] (policyNet)
    {\textbf{Policy network} $\pi_{\boldsymbol{\theta}}$\\[-0.2mm]
     predicts $\mathbf{w}_h,\;w_n,\;\boldsymbol{\phi}$};

\node[optimizer, right=6mm of policyNet, minimum width=27mm] (opt)
    {\textbf{Differentiable}\\ optimizer};

\node[output, right=of opt, minimum width=25mm] (traj)
    {\textbf{Planned control}\\ $\mathbf{u}^{(L)}_{\mathrm{mean}}$};

\node[input, below=7mm of policyNet, minimum width=25mm] (state)
    {\textbf{State, Goal}\\ $\mathbf{x}_0,\;\mathbf{g}$};

\node[fixed, below=7mm of opt, minimum width=31mm] (risk)
    {\textbf{Learned collision model}\\
     $J_{\mathrm{neural}}(\mathbf{o},\mathbf{u})$};

\draw[forward] (images.east) -- (policyNet.west);
\draw[forward] (policyNet.east) -- (opt.west);
\draw[forward] (opt.east) -- (traj.west);

\draw[forward] (state.north) -- (policyNet.south);
\draw[forward] (risk.north) -- (opt.south);

\end{tikzpicture}%
}
\caption{Overview of the model-based imitation-learning policy. 
A frozen depth estimator processes monocular image history. 
The policy network combines visual features, robot state, and goal 
to predict cost weights and optimizer parameters. 
A differentiable optimizer then plans controls using a pre-trained, 
fixed collision model.}
\label{fig:overview_pipeline}
\Description{Image history, robot state, and goal feed a policy network
that predicts cost weights and optimizer parameters. A differentiable optimizer
uses these predictions and a learned collision model to produce planned controls.}
\vspace{-0.3cm}
\end{figure}

\subsection{Overview of the Approach}
\label{sec:policy_pipeline}
\label{diffopt_layer}

\noindent Consider the discrete-time dynamical system $\mathbf{x}_{k+1} =
\mathbf{f}(\mathbf{x}_k, \mathbf{u}_k)$ with initial state $\mathbf{x}_0$. Over a
planning horizon $H$, the control sequence is denoted by $\mathbf{u} = [\mathbf{u}_0,
\ldots, \mathbf{u}_{H-1}] \in \mathbb{R}^{N_u \times H}$, where each control input
satisfies $\mathbf{u}_k \in \mathbb{R}^{N_u}$. The admissible control set is defined as:
\begin{align}
    \mathcal{U}
    = \left\{
        \mathbf{u} \in \mathbb{R}^{N_u \times H}
        \mathrel{\Big|}
        \substack{
            \mathbf{u}_{\min} \le \mathbf{u}_k \le \mathbf{u}_{\max}, \\
            k = 0, \ldots, H-1
        }
    \right\}.
    \label{control_set}
\end{align}
As shown in Fig.~\ref{fig:overview_pipeline}, our local navigation policy at state
$\mathbf{x}_0$ is framed as the following optimization problem:

\begin{equation}
    \mathbf{u}^{\star}
    \in
    \operatorname*{arg\,min}_{\mathbf{u} \in \mathcal{U}}
    \big[
        \mathbf{w}_h^{\top} \mathbf{J}_{\mathrm{hand}}(\mathbf{u})
        + w_n J_{\mathrm{neural}}(\mathbf{o}, \mathbf{u})
    \big].
    \label{risk_aware_optimization}
\end{equation}
The vector $\mathbf{J}_{\mathrm{hand}}(\mathbf{u}) \in \mathbb{R}^{n_h}$ aggregates
handcrafted objectives, such as goal progress, trajectory smoothness, and control
effort. The \emph{model-based} component of our approach learns the neural cost
$J_{\mathrm{neural}}(\mathbf{o}, \mathbf{u})$, which estimates collision risk
conditioned on candidate controls $\mathbf{u}$ and an observation vector $\mathbf{o} =
(I_{0:m}, \mathbf{x}_0)$ comprising a history of monocular image frames $I_{0:m}$ and
the current state $\mathbf{x}_0$.

The \emph{imitation learning} component of our approach infers the weighting parameters
$\mathbf{w}_h$ and $w_n$ from human demonstrations. This enables the policy to balance
goal progress and smoothness against collision risk based on teleoperated robot
trajectories in obstacle-rich environments. In addition to the cost weights, the same
imitation learning framework predicts algorithmic hyperparameters $\mathbf{\phi}$ (e.g.,
a warm start) for the trajectory optimizer, substantially reducing inference latency.

A key challenge is that expert demonstrations provide supervision only in the
form of desired trajectories, rather than target values for the cost weights or
optimizer hyperparameters. We therefore learn these quantities through their
downstream effect on the trajectory produced by the planner. Specifically, we employ a 
policy network (Fig.~\ref{fig:overview_pipeline}) that maps the current navigation
context to cost weights and optimization hyperparameters, which are then passed to a
differentiable optimization layer. The output of this layer is the optimal control sequence
and its induced trajectory is compared with the expert trajectory. The gradients of this
imitation loss are traced through the optimization layer to update the policy network.

In the following section, we first present our approach for learning
$J_{\mathrm{neural}}$, followed by a differentiable-optimization-based imitation-learning
approach.

\subsection{Learned Neural Collision-Risk Cost}
\label{sec:training}

\noindent Let $d(\mathbf{o},\mathbf{u};\boldsymbol{\psi})$ be a learned model
which denotes the worst-case obstacle clearance along a
given control sequence $\mathbf{u}$ and is conditioned on the current
observation $\mathbf{o}$. Let $d_{0}(\mathbf{o};\boldsymbol{\alpha})$ be another learned
model which denotes the worst-case obstacle clearance
at the current position of the robot. The neural collision-risk cost in
\eqref{risk_aware_optimization} is constructed from
$d(\mathbf{o},\mathbf{u};\boldsymbol{\psi})$ and
$d_{0}(\mathbf{o};\boldsymbol{\alpha})$.

\begin{figure*}[!t]
    \centering
    \includegraphics[width=0.9\textwidth]{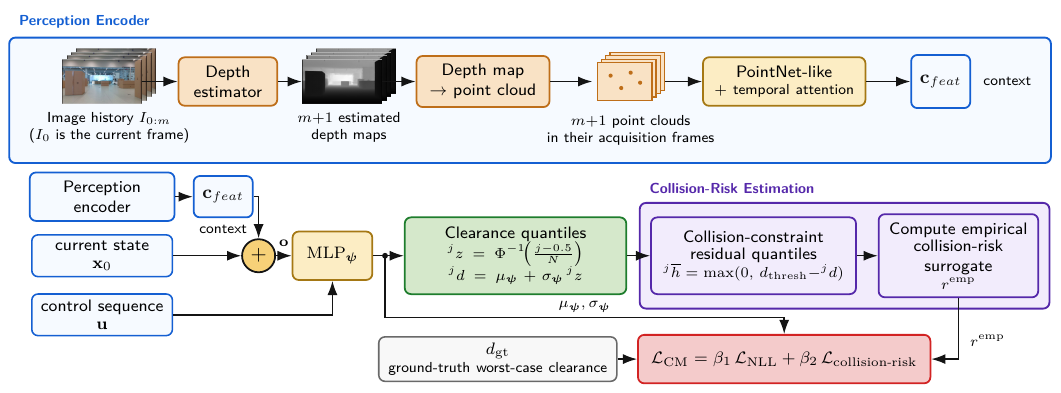}
    \caption{Training pipeline for the history-aware probabilistic collision
    model. The control-dependent clearance branch is shown. 
    Monocular history is converted to point clouds in their 
    acquisition-time robot frames and temporally encoded to predict 
    the worst-case clearance distribution. Fixed midpoint quantiles 
    yield collision-constraint residuals $\overline h$ and the empirical 
    risk surrogate $r^{\mathrm{emp}}$. Gaussian NLL fits the distribution, 
    while one-sided safe/unsafe supervision shapes the risk surrogate.}
    \label{fig:collision_model_pipeline}
    \Description{Collision-model training pipeline. Image history is encoded
    as temporal point-cloud context and combined with robot state and a
    control sequence to predict a clearance distribution. Deterministic
    quantiles produce a collision-risk surrogate, supervised together with
    the clearance likelihood.}
\end{figure*}

\subsubsection{DCBF-Inspired Collision-Risk Cost}
\label{sec:cbf_neural}
\label{sec:chance_risk}
A potential collision avoidance constraint using the learned clearance model
$d(\mathbf{o},\mathbf{u};\boldsymbol{\psi})$ can be formulated as \cite{11283048}:
\begin{align}
    h\big(d(\mathbf{o},\mathbf{u};\boldsymbol{\psi})\big)
    &=-d(\mathbf{o},\mathbf{u};\boldsymbol{\psi})+d_{\mathrm{thresh}}\leq0.
    \label{collision_constraint}
\end{align}
The above constraint is violated when the worst-case obstacle clearance
falls below $d_{\mathrm{thresh}}$. As a result, a
fast-moving robot can satisfy \eqref{collision_constraint} while approaching
an obstacle with its predicted clearance only slightly above the threshold,
leaving insufficient room to brake in the next planning cycle. We therefore
augment \eqref{collision_constraint} with a barrier-based condition
inspired by discrete control barrier functions
\cite{agrawal2017discrete,9483029}. Define
\begingroup
\begin{equation}
\begin{aligned}
    h_0(\mathbf{o};\boldsymbol{\alpha})
    &=-d_0(\mathbf{o};\boldsymbol{\alpha})+d_{\mathrm{thresh}},\\
    h_1(\mathbf{o},\mathbf{u};\boldsymbol{\psi})
    &=-d(\mathbf{o},\mathbf{u};\boldsymbol{\psi})+d_{\mathrm{thresh}}.
\end{aligned}
\label{collision_constraint_evaluations}
\end{equation}
\endgroup
The barrier-based collision avoidance constraint is
\begin{equation}
\begin{aligned}
    h_b(\mathbf{o},\mathbf{u};\boldsymbol{\psi},\boldsymbol{\alpha})
    &\triangleq h_1(\mathbf{o},\mathbf{u};\boldsymbol{\psi})\\
    &\quad -(1-\gamma)h_0(\mathbf{o};\boldsymbol{\alpha})\leq 0,
    \quad\gamma\in(0,1].
\end{aligned}
\label{horizon_barrier}
\end{equation}
The parameter $\gamma$ controls how fast the predicted clearance is allowed to decrease
relative to its current value. A smaller $\gamma$ increases conservativeness and makes
the robot react to obstacles from a larger distance. Larger values permit a greater
decrease in clearance, while
$\gamma=1$ recovers the original collision-avoidance constraint in
\eqref{collision_constraint}.


\noindent \textbf{Extension to Probabilistic Setting:} Rather than predicting a single
value for
$d(\mathbf{o},\mathbf{u};\boldsymbol{\psi})$ and
$d_0(\mathbf{o};\boldsymbol{\alpha})$, it is more useful to predict a distribution that
captures uncertainty arising from
noisy monocular depth estimates
\cite{depthanything,NEURIPS2024_26cfdcd8,https://doi.org/10.48550/arxiv.2302.12288}. We
model $d(\mathbf{o},\mathbf{u};\boldsymbol{\psi})$ and
$d_0(\mathbf{o};\boldsymbol{\alpha})$ as
\begingroup
\begin{equation}
\begin{aligned}
    d(\mathbf{o},\mathbf{u};\boldsymbol{\psi})
    &\sim\mathcal{N}\big(
        \mu_{\boldsymbol{\psi}}(\mathbf{o},\mathbf{u}),
        \sigma_{\boldsymbol{\psi}}^2(\mathbf{o},\mathbf{u})
    \big),\\
    d_0(\mathbf{o};\boldsymbol{\alpha})
    &\sim\mathcal{N}\big(
        \mu_{0,\boldsymbol{\alpha}}(\mathbf{o}),
        \sigma_{0,\boldsymbol{\alpha}}^2(\mathbf{o})
    \big).
\end{aligned}
\label{clearance_distributions}
\end{equation}
\endgroup

The predicted variances quantify uncertainty in the corresponding clearance
values. The distributions of $d$ and $d_0$ induce a Gaussian distribution
$\mathcal{N}(\mu_{h_b}, \sigma^2_{h_b})$
over $h_b$ (see the supplementary text). Consequently,
deterministic enforcement of the collision-avoidance constraint
\eqref{horizon_barrier} is no longer appropriate. Instead, it is expressed as the chance
constraint
$\mathbb{P}\!\left(
        h_b(\mathbf{o},\mathbf{u};\boldsymbol{\psi},\boldsymbol{\alpha})\leq0
    \right)
    \geq 1-\varepsilon$ or, equivalently,
$\mathbb{P}(h_b(\mathbf{o},\mathbf{u};\boldsymbol{\psi},\boldsymbol{\alpha})>0)\leq\varepsilon$.
We define the associated collision risk as
\begin{align}
    r\!\left(h_b(\mathbf{o},\mathbf{u};\boldsymbol{\psi},\boldsymbol{\alpha})\right)
    &\triangleq
    \mathbb{P}\!\left(
        h_b(\mathbf{o},\mathbf{u};\boldsymbol{\psi},\boldsymbol{\alpha})>0
    \right).
    \label{risk_def}
\end{align}
The collision risk is the probability that a control sequence violates the
collision-avoidance constraint in \eqref{horizon_barrier}. A larger value
indicates that the predicted clearance is more likely to decrease faster than
the constraint permits. A smaller value indicates that the control sequence is
more likely to satisfy the constraint.

Rather than computing the collision risk in \eqref{risk_def} analytically, we
approximate it using $r^{\mathrm{emp}}$, an empirical collision-risk
surrogate. This surrogate can be evaluated using either random samples or
fixed deterministic Gaussian quantiles drawn from $\mathcal{N}(\mu_{h_b},
\sigma^2_{h_b})$. We use deterministic Gaussian
quantiles instead of Monte Carlo samples to avoid sampling variability. Thus,
fixed observation and control inputs $(\mathbf{o},\mathbf{u})$ always produce
the same $r^{\mathrm{emp}}$ value. The supplementary text
provides the quantile construction.

The algebraic form of $r^{\mathrm{emp}}$ depends on what kind of risk metric we want to
evaluate. Common choices include sample average approximation (SAA)
\cite{pagnoncelli2009sample,Shirai_2023,shapiro2014lectures} and conditional
value-at-risk (CVaR) \cite{yin2023risk,DBLP:journals/ral/LewBP24}. SAA computes the
empirical violation frequency while CVaR aggregates the upper-tail constraint
values. Our method is agnostic to
the choice of $r^{\mathrm{emp}}$. In this work, we
instantiate $r^{\mathrm{emp}}$ using Maximum Mean Discrepancy (MMD)
\cite{10.5555/2188385.2188410,11283048,11084873,10979910},
whose construction is detailed in
the supplementary text.

The neural collision-risk cost used in \eqref{risk_aware_optimization} is
\begin{align}
    J_{\mathrm{neural}}(
        \mathbf{o},\mathbf{u};\boldsymbol{\psi},\boldsymbol{\alpha}
    )
    &=\left[
        r^{\mathrm{emp}}\!\left(
            \{{}^{j}\overline{h}_b(
                \mathbf{o},\mathbf{u};\boldsymbol{\psi},\boldsymbol{\alpha}
            )\}_{j=1}^{N}
        \right)-\tau
    \right]_+.
    \label{barrier_neural_cost}
\end{align}
where ${}^{j}\overline{h}_b = \max(0, {}^{j}{h}_b)$ and ${}^{j}{h}_b$ is the $j^{th}$
Gaussian quantile derived from $\mathcal{N}(\mu_{h_b}, \sigma^2_{h_b})$ (see the supplementary text).
Furthermore, $[a]_+=\max(0,a)$, and $\tau$ is the threshold applied to the empirical
collision-risk surrogate. The value of $\tau$ is reported in
Section~\ref{sec:implementation}. Equation~\eqref{barrier_neural_cost} is a soft
penalty-based surrogate for enforcing the chance constraint
$r\leq\varepsilon$. A restrictive
$\varepsilon$ can reject useful motion under predictive uncertainty and
lead to overly conservative behavior. Thus, the weighted neural cost
allows the optimizer to trade surrogate-risk exceedance against task
progress. 

\subsubsection{Task-Aware Collision-Model Training}
\label{sec:collision_training}

Having constructed the neural collision-risk cost in
Section~\ref{sec:cbf_neural}, we now present how the collision model
$d(\mathbf{o},\mathbf{u};\boldsymbol{\psi})$ is learned. As summarized in
Figure~\ref{fig:collision_model_pipeline}, this model maps the visual
observation history, current robot state, and a given control sequence to the
distribution of worst-case obstacle clearance along that sequence. For each
observation $\mathbf{o}$ and control sequence $\mathbf{u}$, we roll out the
corresponding trajectory and use privileged LiDAR measurements to collect its
ground-truth worst-case obstacle clearance $d_{\mathrm{gt}}$.
The collision-model dataset contains both safe and unsafe motion primitives,
where $d_{\mathrm{gt}}\geq d_{\mathrm{thresh}}$ denotes a safe primitive and
$d_{\mathrm{gt}}<d_{\mathrm{thresh}}$ denotes an unsafe primitive.

Following the task-aware collision-model training framework of
MonoMPC~\cite{11283048}, we combine distribution fitting with task-aware
supervision. A standard way to train a model that outputs a Gaussian
distribution is to minimize the Gaussian negative log-likelihood (NLL):
\begin{align}
    \mathcal{L}_{\mathrm{NLL}}
    &=\frac{1}{2}\log\!\left(
        2\pi\sigma_{\boldsymbol{\psi}}^2(\mathbf{o},\mathbf{u})
      \right)
      +\frac{\left(
        d_{\mathrm{gt}}-\mu_{\boldsymbol{\psi}}(\mathbf{o},\mathbf{u})
      \right)^2}
      {2\sigma_{\boldsymbol{\psi}}^2(\mathbf{o},\mathbf{u})}.
    \label{gaussian_nll_loss}
\end{align}

However, NLL alone does not supervise how the variance should behave in downstream
planning. A safe trajectory may be assigned high collision risk because of an
inflated variance, or an unsafe trajectory may be assigned low collision risk
because of an underestimated variance. We therefore augment the NLL loss with
the collision-risk estimation head shown in
Figure~\ref{fig:collision_model_pipeline}, which provides the task-aware supervision.
During collision-model training, the residuals are computed directly from
raw clearance, without the barrier formulation:
${}^{j}\overline h=\max(0,d_{\mathrm{thresh}}-{}^{j}d)$, where
${}^{j}d=\mu_{\boldsymbol{\psi}}(\mathbf{o},\mathbf{u})
+\sigma_{\boldsymbol{\psi}}(\mathbf{o},\mathbf{u})\,{}^{j}z$ and
${}^{j}z=\Phi^{-1}((j-0.5)/N)$ for $j=1,\ldots,N$.
Here $\Phi^{-1}$ is the inverse standard-normal cumulative distribution
function. Specifically,
\begin{align}
    \mathcal{L}_{\mathrm{safe}}
    &=\left[
        r^{\mathrm{emp}}\!\left(
            \{{}^{j}\overline h(
                \mathbf{o},\mathbf{u};\boldsymbol{\psi}
            )\}_{j=1}^{N}
        \right)-\tau
      \right]_+^2,
    \label{risk_loss_safe}\\
    \mathcal{L}_{\mathrm{unsafe}}
    &=\left[
        r^{\star}-r^{\mathrm{emp}}\!\left(
            \{{}^{j}\overline h(
                \mathbf{o},\mathbf{u};\boldsymbol{\psi}
            )\}_{j=1}^{N}
        \right)
      \right]_+^2,
    \label{risk_loss_unsafe}\\
    \mathcal{L}_{\mathrm{collision\text{-}risk}}
    &=\begin{cases}
        \omega_{\mathrm{safe}}\mathcal{L}_{\mathrm{safe}},
        &d_{\mathrm{gt}}\geq d_{\mathrm{thresh}},\\
        \omega_{\mathrm{unsafe}}\mathcal{L}_{\mathrm{unsafe}},
        &d_{\mathrm{gt}}<d_{\mathrm{thresh}},
      \end{cases}
    \label{collision_risk_loss}
\end{align}
The total loss is $\mathcal{L}_{\mathrm{CM}}=\beta_1\mathcal{L}_{\mathrm{NLL}}
+\beta_2\mathcal{L}_{\mathrm{collision\text{-}risk}}$.
Equation~\eqref{risk_loss_safe} penalizes a safe primitive if its
predicted collision-risk surrogate exceeds the threshold $\tau$. Thus, variance is
regularized in a way that the risk-surrogate values of safe primitives remain below the
prescribed threshold. In contrast, equation~\eqref{risk_loss_unsafe} penalizes an unsafe
primitive when
its predicted surrogate is below the target
$r^{\star}=\max\!\left(\zeta s+(1-\zeta)s^2,r_{\mathrm{floor}}\right)$.
Here $s=(d_{\mathrm{thresh}}-d_{\mathrm{gt}})/d_{\mathrm{thresh}}$ is its
normalized penetration severity. The parameter $\zeta$ blends the linear and
quadratic dependence of $r^{\star}$ on this severity, while
$r_{\mathrm{floor}}$ ensures a minimum target for every unsafe primitive.

Equation~\eqref{collision_risk_loss} selects the appropriate one-sided loss
using the ground-truth safe or unsafe label. The coefficients
$\omega_{\mathrm{safe}}$ and $\omega_{\mathrm{unsafe}}$ control the relative
influence of the two classes. Finally, $\mathcal{L}_{\mathrm{CM}}$ combines
Gaussian distribution fitting through $\mathcal{L}_{\mathrm{NLL}}$ with the
collision-risk supervision through
$\mathcal{L}_{\mathrm{collision\text{-}risk}}$. The coefficients $\beta_1$
and $\beta_2$ balance these two objectives and are specified in
Section~\ref{sec:implementation}.
The current worst-case obstacle clearance model $d_0(\mathbf{o};\boldsymbol{\alpha})$ is
trained
similarly to $d(\mathbf{o},\mathbf{u};\boldsymbol{\psi})$.

\subsubsection{Neural Network Architecture}
\label{sec:collision_architecture}

Figure~\ref{fig:collision_model_pipeline} summarizes the colli\-sion-model
architecture for $d(\mathbf{o},\mathbf{u};\boldsymbol{\psi})$. Each of the past $m = 20$
image frames and the current one are passed to a frozen
Depth Anything V2~\cite{NEURIPS2024_26cfdcd8} to obtain metric depth estimates. Each
depth map is then back-projected to
a two-dimensional point cloud. A shared PointNet-like encoder~\cite{qi2017pointnet}
independently maps each
point cloud to a frame-level feature. Temporal self-attention then aggregates
these features into the context $\mathbf{c}_{\mathrm{feat}}$. We concatenate
$\mathbf{c}_{\mathrm{feat}}$ with current state $\mathbf{x}_0$ and control sequence
$\mathbf{u}$ and pass them through a multi-layer perceptron $\mathbf{MLP}_\psi$ to
obtain the mean and variance of the worst-case obstacle clearance distribution. A
separate risk-head then uses these predictions to evaluate $r^{\mathrm{emp}}$ and the
task-risk loss discussed in the previous subsection. For our MMD-based
$r^{\mathrm{emp}}$, the $\mathbf{MLP}_\psi$ also predicts the control-dependent kernel
width
$\lambda_{\boldsymbol{\psi}}(\mathbf{o},\mathbf{u})$, which is used to compute $r^{\mathrm{emp}}$
as detailed in
the supplementary text.

\begin{algorithm}[!t]
\caption{Soft-CEM$(\mathbf{w}_h,w_n,\boldsymbol{\phi};n,L)$}
\label{algo_1}
\DontPrintSemicolon
\KwIn{cost weights $\mathbf{w}_h,w_n$ in
\eqref{risk_aware_optimization}, learned hyperparameters
$\boldsymbol{\phi}=(\mathbf{u}_{\mathrm{mean}}^{(0)},
\boldsymbol{\rho}^{(0)},\eta)$, control set $\mathcal U$,
stochastic candidate count $n$, refinements $L$}
\KwOut{final mean control sequence $\mathbf{u}_{\mathrm{mean}}^{(L)}$}
\BlankLine
Initialize the sampling mean, element-wise variance, and temperature
from $\boldsymbol{\phi}$, with $\mathbf{u}_{\mathrm{mean}}^{(0)}\in\mathcal U$
\quad\mbox{\eqref{policy_outputs}}\;
\For{$l\gets1$ \KwTo $L$}{
    Form candidates from $n$ projected samples \eqref{cem_training_samples}
    and $\mathbf{u}_{\mathrm{mean}}^{(l-1)}$, indexed by
    $\mathcal Q=\{1,\ldots,n+1\}$\;
    $J_q^{(l)}\gets
    \mathbf{w}_h^{\top}\mathbf{J}_{\mathrm{hand}}(\mathbf{u}_q^{(l)})
    +w_nJ_{\mathrm{neural}}(\mathbf{o},\mathbf{u}_q^{(l)})$,
    $q\in\mathcal{Q}$
    \quad\mbox{\eqref{risk_aware_optimization}}\;
    Compute soft weights
    $\{\alpha_q^{(l)}\}_{q\in\mathcal{Q}}$ using the fixed temperature $\eta$
    \quad\mbox{\eqref{cem_training_weights}}\;
    Update $\mathbf{u}_{\mathrm{mean}}^{(l)}$ and
    $\boldsymbol{\rho}^{(l)}$ by weighted moments, clipping
    the variance to $[\rho_{\min},\rho_{\max}]$
    \quad\mbox{\eqref{cem_training_update}}\;
}
\Return $\mathbf{u}_{\mathrm{mean}}^{(L)}$
\label{ln:cem_return}\;
\end{algorithm}

\subsection{Differentiable Planning and Policy Network Learning}
\label{sec:planning_algorithm}

\noindent The optimizer in \eqref{risk_aware_optimization} is realized by soft-CEM, a
differentiable variant of conventional CEM~\cite{botev2013cross}, motivated by
Amos and Yarats~\cite{pmlr-v119-amos20a}. The overall algorithm is shown in
Alg.~\ref{algo_1}. Assume we have access to a policy network
$\boldsymbol{\pi}_{\boldsymbol{\theta}}$ that provides the following predictions

\begin{equation}
\begin{aligned}
    (\boldsymbol{w}_h,w_n,\boldsymbol{\phi})
    &=\pi_{\boldsymbol{\theta}}(I_{0:m},\mathbf{x}_0,\mathbf{g}),\\
    \boldsymbol{\phi}&=(\mathbf{u}_{\mathrm{mean}}^{(0)},
    \boldsymbol{\rho}^{(0)},\eta)\in\mathbb{R}^{n_{\phi}}.
\end{aligned}
\label{policy_outputs}
\end{equation}

\noindent Here $\mathbf{u}_{\mathrm{mean}}^{(0)}$ and $\boldsymbol{\rho}^{(0)}$ are the mean
and element-wise variance, respectively, used to initialize the CEM algorithm in Alg.~\ref{algo_1}. The
initial mean lies in $\mathcal U$. The
scalar $\eta$ is the soft-weighting temperature and controls how strongly refinement
emphasizes
lower-cost candidates. The weights $\boldsymbol{w}_h,w_n$ are fed to the cost function
in \eqref{risk_aware_optimization} to trade off collision risk with other task-relevant
costs.

\noindent \textbf{Control Sampling:} At iteration $l$ of our CEM, Alg.~\ref{algo_1}, we
first sample $n$ control input sequences using the reparameterization trick in the
following manner.

\begin{equation}
    \mathbf{u}_q^{(l)}
    =\Pi_{\mathcal{U}}\!\left(
        \mathbf{u}_{\mathrm{mean}}^{(l-1)}
        +\sqrt{\boldsymbol{\rho}^{(l-1)}}
        \odot\boldsymbol{\xi}_q^{(l)}
    \right),
    \quad q=1,\ldots,n,
    \label{cem_training_samples}
\end{equation}
\noindent where $\boldsymbol{\xi}_q^{(l)}\sim \mathcal{N}(\mathbf{0},\mathbf{I})$,
$\Pi_{\mathcal{U}}$ projects $\mathbf{u}_q$ to $\mathcal{U}$. We also append the
previous mean as an additional candidate,
$\mathbf{u}_{n+1}^{(l)}
=\mathbf{u}_{\mathrm{mean}}^{(l-1)}$,
giving $n+1$ candidate control sequences indexed by
$\mathcal{Q}=\{1,\ldots,n+1\}$.

\noindent\textbf{Soft weighting.} Each candidate is scored by
$J_q^{(l)}$ using
\eqref{risk_aware_optimization}, with the frozen risk model providing the collision-risk
part \eqref{barrier_neural_cost} of the cost. Writing
$J_{\min}^{(l)}=\min_{q'\in\mathcal{Q}}J_{q'}^{(l)}$,
\begin{align}
    \alpha_q^{(l)}
    &=\frac{
        \exp\!\left(
            -(J_q^{(l)}-J_{\min}^{(l)})/\eta
        \right)
    }{
        \sum_{q'\in\mathcal{Q}}
        \exp\!\left(
            -(J_{q'}^{(l)}-J_{\min}^{(l)})/\eta
        \right)
    }.
    \label{cem_training_weights}
\end{align}
The temperature $\eta$ sets how strongly low-cost candidates dominate, and
subtracting $J_{\min}^{(l)}$ is a numerical stabilization that leaves
$\alpha^{(l)}$ unchanged.

\begin{figure*}[!t]
    \centering
    \includegraphics[width=0.9\textwidth]{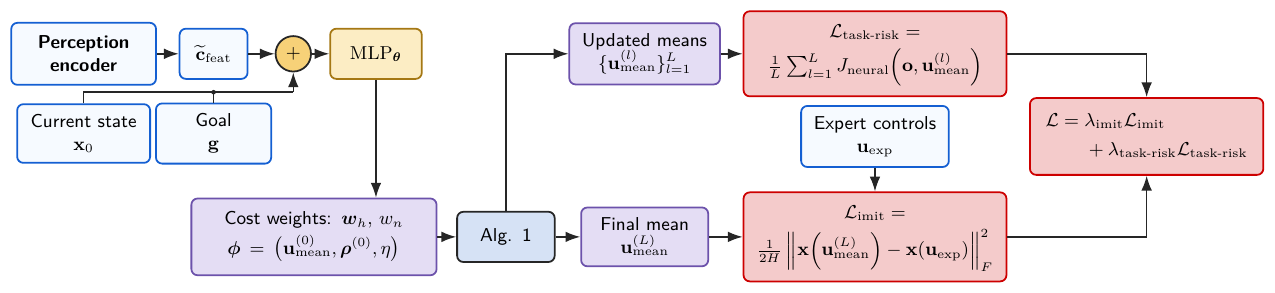}
    \caption{Policy-network training through soft-CEM. The perception encoder is an
    independent copy of the architecture in
    Fig.~\ref{fig:collision_model_pipeline}, with separate parameters.
    Its context $\widetilde{\mathbf c}_{\mathrm{feat}}$, the current state
    $\mathbf x_0$, and goal $\mathbf g$ determine the cost weights
    $\boldsymbol w_h,w_n$ and optimizer hyperparameters
    $\boldsymbol{\phi}=(\mathbf u_{\mathrm{mean}}^{(0)},
    \boldsymbol{\rho}^{(0)},\eta)$.
    The Alg.~\ref{algo_1} box represents the complete soft-CEM procedure
    with $L$ refinement iterations.
    The updated means define $\mathcal L_{\mathrm{task\text{-}risk}}$.
    The final mean's induced planar trajectory is compared with the expert
    trajectory through $\mathcal L_{\mathrm{imit}}$.
    Gradients of the weighted loss pass through Alg.~\ref{algo_1} to train
    the policy network while the clearance-model parameters remain fixed.}
    \label{fig:diffcem_pipeline}
    \Description{Policy-training pipeline. A perception encoder, robot state,
    and goal determine cost weights and optimizer hyperparameters.
    Algorithm 1 returns updated control means and a final mean. The
    intermediate-risk and final trajectory-imitation losses are combined
    to train the policy network.}
\end{figure*}

\noindent\textbf{Moment update.} The mean and variance of the control sampling
distribution are updated using $\alpha_q^{(l)}$ in the following manner.

\begingroup
\begin{equation}
\begin{aligned}
    \mathbf{u}_{\mathrm{mean}}^{(l)}
    &=\sum_{q\in\mathcal Q}\alpha_q^{(l)}\mathbf{u}_q^{(l)},\\
    \boldsymbol{\rho}^{(l)}
    &=\Pi_{[\rho_{\min},\rho_{\max}]}\Big(
        \sum_{q\in\mathcal Q}\alpha_q^{(l)}
        \big(\mathbf{u}_q^{(l)}-\mathbf{u}_{\mathrm{mean}}^{(l)}\big)^{\odot2}
    \Big),
\end{aligned}
\label{cem_training_update}
\end{equation}
\endgroup
where $(\cdot)^{\odot2}$ is element-wise squaring and the variance projection
acts element-wise. Each updated mean satisfies the control bounds by construction,
being a convex combination of admissible control sequences. The temperature $\eta$ 
remains constant across CEM
iterations. It should be noted that unlike hard-elite CEM, every control sample
contributes to the moment update.

Re-parameterized sampling, cost evaluation, soft weighting, and the moment
refit are differentiable, while the projections are differentiable almost
everywhere. Since both the cost weights and $\boldsymbol{\phi}$ depend on
$\boldsymbol{\theta}$, the derivative
$\partial\mathbf{u}_{\mathrm{mean}}^{(L)}/\partial\boldsymbol{\theta}$ is
computed by differentiating through all $L$ unrolled iterations.

\noindent \textbf{Learning the Policy Network:} Figure~\ref{fig:diffcem_pipeline}
summarizes policy network training. The core network
architecture consists of
an independent copy of the perception architecture in
Section~\ref{sec:collision_architecture}, with separate parameters, to
produce $\widetilde{\mathbf{c}}_{\mathrm{feat}}$. This context is
combined with $\mathbf{x}_0$ and $\mathbf{g}$ and passed through an MLP to obtain
$\boldsymbol{w}_h,w_n,\boldsymbol{\phi}$, which are then passed to the soft-CEM block of
Alg.~\ref{algo_1}.

Let $\mathbf{u}_{\mathrm{mean}}^{(L)}$ be the control sequence obtained by running $L$
CEM refinement iterations and let
$\mathbf{x}(\mathbf{u}_{\mathrm{mean}}^{(L)})\in\mathbb{R}^{2\times H}$ be its induced
planar trajectory. The final mean is supervised against the expert trajectory:
\begin{align}
    \mathcal{L}_{\mathrm{imit}}
    &=\frac{1}{2H}
      \left\|
        \mathbf{x}(\mathbf{u}_{\mathrm{mean}}^{(L)})
        -\mathbf{x}(\mathbf{u}_{\mathrm{exp}})
      \right\|_F^2.
    \label{policy_imitation_loss}
\end{align}
The subscript $F$ denotes the Frobenius norm, and the normalization
averages over the $2H$ planar trajectory coordinates. We also accumulate $J_{\mathrm{neural}}$
evaluated at every updated control mean and include that as the additional task loss
for the policy network training.
\begin{align}
    \mathcal{L}_{\mathrm{task\text{-}risk}}
    &=\frac{1}{L}\sum_{l=1}^{L}
      J_{\mathrm{neural}}\!\left(
        \mathbf{o},\mathbf{u}_{\mathrm{mean}}^{(l)}
      \right).
    \label{policy_risk_loss}
\end{align}
The total loss is $\mathcal{L}
=\lambda_{\mathrm{imit}}\mathcal{L}_{\mathrm{imit}}
+\lambda_{\mathrm{task\text{-}risk}}\mathcal{L}_{\mathrm{task\text{-}risk}}$.
The imitation term brings the CEM output trajectory closer to the expert demonstrations,
while the second term improves safety of each intermediate solution. We can
backpropagate $\mathcal L$ through the unrolled Alg.~\ref{algo_1} iterations to train
the policy network to produce scene-dependent cost weights and algorithmic
hyperparameters.


\section{Validation and Benchmarking}\label{sec:validation}

We evaluate navigation safety, efficiency, steering smoothness, and 
computation time in real-world experiments. Metric definitions, aggregation rules, 
and timing details are provided in the supplementary text.

\begin{figure*}[!t]
    \centering
    \includegraphics[width=0.95\textwidth]{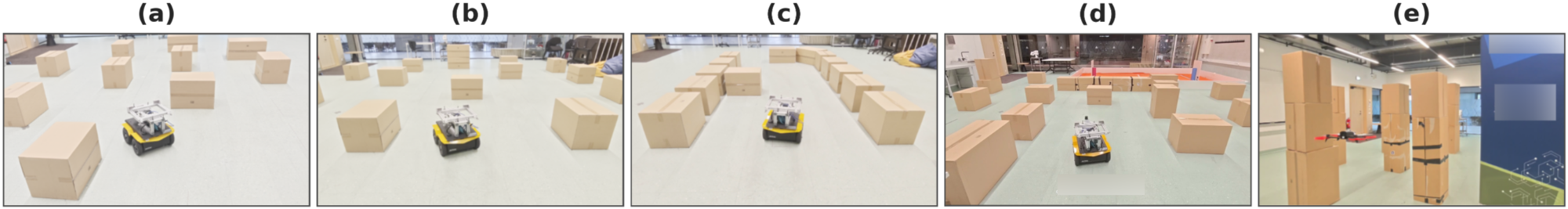}
    \caption{{Panels (a)–(c) show the Grid-Difficult, Grid-Easy, and
    Corridor
    environments, respectively, used to evaluate our method and the baselines on the
    Jackal.
    Panel (d) shows the more challenging grid environment used for ablations A1 and A2. 
    Panel (e) shows the environment used for
    the Parrot Bebop experiments. }}
    \label{fig:real_world_snapshots}
    \Description{Snapshots of the Grid-Difficult, Grid-Easy, and Corridor
    environments used for Jackal experiments, the more challenging grid
    environment used for fixed-initialization ablations, and the environment
    used for Parrot Bebop experiments.}
\end{figure*}

\subsection{Experimental and Implementation Setup}
\label{sec:implementation}
\noindent\textbf{Benchmarking setup:}
We evaluate on a Clearpath Jackal using only the RGB stream of an Intel
RealSense D435i camera ($69^\circ$ horizontal FOV). A SICK TIM551 2D LiDAR
is used solely for ground-truth evaluation and visualization. The system
runs ROS Noetic on a locally networked RTX~3080 laptop, with
TensorRT-accelerated Depth Anything V2~\cite{NEURIPS2024_26cfdcd8}.
Tests use static cardboard-box obstacles in Corridor, Grid-Easy, and
Grid-Difficult environments. Figure~\ref{fig:real_world_snapshots} shows
the Jackal and Parrot Bebop trial environments. The accompanying video
provides further details.

\noindent\textbf{Planning setup:}
Alg.~\ref{algo_1} is implemented in JAX~\cite{jax2018github} and executed
in a receding-horizon MPC loop. With $n_h=1$, the vector $\boldsymbol{w}_h$ reduces
to the scalar $w_h$, and the hand-crafted cost in \eqref{risk_aware_optimization} is
$J_{\mathrm{hand}}(\mathbf{u})
=\sum_{k=0}^{H-1}[(x_k-x_f)^2+(y_k-y_f)^2]$,
where $(x_k,y_k)$ is the position component of the rolled-out state
$\mathbf{x}_{k+1}$ and $(x_f,y_f)$ is the goal. Each sequence contains
$H=50$ linear- and angular-velocity pairs over $5$\,s
($\Delta t=0.1$\,s), bounded by $[0,1]$\,m/s and $[-1,1]$\,rad/s,
respectively. Initial sampling variances are bounded elementwise by
$[\rho_{\min},\rho_{\max}]=[10^{-4},5]$. Softplus keeps $w_h$, $w_n$,
and $\eta$ positive.

\noindent\textbf{Shared dataset:}
The collision model and policy share $278{,}518$ scenes collected through
teleoperation and a privileged LiDAR-based planner~\cite{Andrejev_2025}.
Each scene contains 21 point clouds derived from $I_{0:m}$ ($m=20$),
covering $2$\,s at $0.1$\,s intervals. Each cloud contains 300
two-dimensional points estimated using Depth Anything
V2~\cite{NEURIPS2024_26cfdcd8}, expressed in the robot frame at acquisition.
Policy-training records are
$(I_{0:m},\mathbf{x}_0,\mathbf{g},\mathbf{u}_{\mathrm{exp}})$,
containing the current state, goal, and 50-step expert control sequence.
The current position is the origin of the ego-centered planning frame,
so only current velocity is stored for $\mathbf{x}_0$.
Each scene also records the expert primitive's ground-truth worst-case
clearance. Following~\cite{11283048}, clearance for a primitive
$\mathbf{u}$ is computed as
\begin{align}
    d_{\mathrm{gt}}(\mathbf{u})
    &=
    \min_{\substack{
        k=0,\ldots,H-1\\
        \mathbf{p}\in\mathcal{P}_{\mathrm{LiDAR}}
    }}
    \left\|
        \begin{bmatrix}
            x_k(\mathbf{u})\\
            y_k(\mathbf{u})
        \end{bmatrix}
        -\mathbf{p}
    \right\|_2,
    \label{ground_truth_clearance}
\end{align}
where $(x_k(\mathbf{u}),y_k(\mathbf{u}))$ is the robot position at
trajectory index $k$ and $\mathcal{P}_{\mathrm{LiDAR}}$ is the privileged
LiDAR cloud. Clearance is measured from the robot center while
$d_{\mathrm{thresh}}=0.35$\,m accounts for the robot radius and a safety
buffer.

\noindent\textbf{Collision-model augmentation:}
For collision-model training, each scene is paired with the expert
primitive and 50 generated motion primitives.
The expert and generated motion primitives are labeled safe/unsafe using
\eqref{ground_truth_clearance}, and their clearance values are recorded.


\begin{figure}[!t]
    \centering
    \includegraphics[width=\columnwidth]{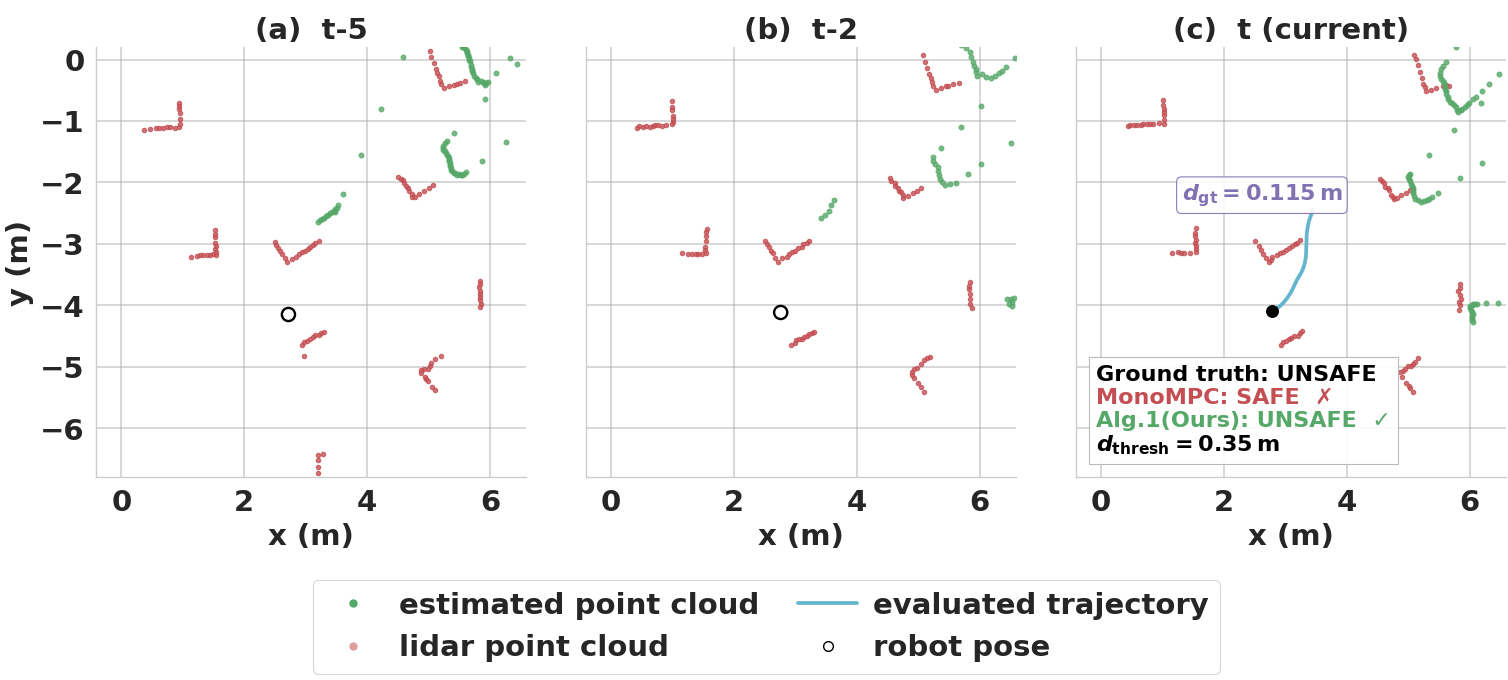}
    \caption{{Comparison of safety predictions using current-frame
    observations and image history. Estimated point clouds and ground-truth
    LiDAR clouds illustrate obstacles near the robot and an unsafe motion
    primitive. Here $t$ denotes the current frame, while $t-2$ and $t-5$
    denote previous frames.} }
    \label{fig:history_effect}
    \Description{Current and past camera observations illustrate an obstacle
    that is outside the current field of view. Estimated point clouds and
    ground-truth LiDAR geometry show a candidate motion primitive passing
    close to the obstacle.}
    \vspace{-0.3cm}
\end{figure}

\noindent\textbf{Training setup:}
Both collision model and policy networks are implemented and trained in
Equinox~\cite{kidger2021equinox} on an RTX~5090.
Collision-model training uses $N=20$ deterministic clearance quantiles
and $d_{\mathrm{thresh}}=0.35$\,m. All experiments use the squared-MMD
surrogate $r^{\mathrm{emp}}$ defined in the supplementary text.
We set $\beta_1=1$, $\beta_2=10$, $\omega_{\mathrm{safe}}=10$,
$\omega_{\mathrm{unsafe}}=1$, $\tau=0.005$, $\zeta=0.5$, and
$r_{\mathrm{floor}}=2\tau=0.01$.
Policy training uses $L=5$ refinements, each with $n=20$ stochastic
candidates and one current-mean anchor, with $\gamma=0.95$,
$\lambda_{\mathrm{imit}}=1$, and $\lambda_{\mathrm{task\text{-}risk}}=10$.

\subsection{Comparison with MonoMPC \texorpdfstring{\cite{11283048}}{}}\label{sec:monompc}

\noindent In this subsection, we compare our approach with a closely related baseline MonoMPC~\cite{11283048} that also uses a learned collision model with a risk-aware trajectory optimization for monocular navigation. This benchmark is particularly important because our approach improves upon MonoMPC~\cite{11283048} by incorporating observation history into collision prediction, improving task-aware training and making the planner adaptable to the complexity of the encountered navigation scenario. We first examine the collision-model predictions to illustrate how temporal context helps address the limited spatial coverage of monocular observations.

\noindent \textbf{Effect of Temporal Context on Collision Risk Prediction:} A nearby obstacle may leave the camera’s field of view while remaining relevant to the safety of a candidate trajectory. Fig.~\ref{fig:history_effect} illustrates this situation through a sequence of past observations and the current frame. The red points show the ground-truth LiDAR geometry, while the green points show the point clouds estimated from monocular images. In Fig.~\ref{fig:history_effect}, we show a trajectory primitive resulting from a control sequence whose safety needs to be assessed. It can be seen that it passes very close to an obstacle (red point cloud)--ground-truth minimum clearance of $d_{\mathrm{gt}}=0.115\,\mathrm{m}$, below the prescribed safety threshold of $d_{\mathrm{thresh}}=0.35\,\mathrm{m}$. However, due to the limited FOV, the robot does not see the obstacle in its current frame. As a result, MonoMPC, which predicts collision risk using only the current image, classifies the primitive as safe.

In contrast, our proposed collision model correctly classifies the primitive as unsafe as it is able to leverage past observations through the perception encoder described in Section~\ref{sec:collision_architecture} to predict the correct distribution of the worst-case clearance. Thus, this example illustrates how temporal context provides a learned memory for collision assessment without constructing an explicit occupancy map.

\begin{table*}[t]
\centering
\caption{Comparison with MonoMPC across three
    environments (30 different runs in each environment per method).}
\label{table_monompc}
\begin{tabular}{llccccc}
    \toprule
    Environment & Method & Collision & \multicolumn{2}{c}{Linear speed (m/s)}
    & \multicolumn{2}{c}{Angular speed (rad/s)} \\
    \cmidrule(lr){4-5}\cmidrule(lr){6-7}
    & & rate (\%) & Avg. & Mean peak & Avg. & Mean peak \\
    \midrule
    \multirow{2}{*}{Corridor} & Alg.~\ref{algo_1} (Ours)
    & \textbf{7} & \textbf{0.77} & \textbf{1.00} & \textbf{0.26} & \textbf{0.82} \\
    & MonoMPC~\cite{11283048} & 37 & 0.28 & 0.86 & 0.39 & 0.93 \\
    \multirow{2}{*}{Grid-Easy} & Alg.~\ref{algo_1} (Ours)
    & \textbf{3} & \textbf{0.81} & \textbf{1.00} & \textbf{0.26} & \textbf{0.78} \\
    & MonoMPC~\cite{11283048} & 23 & 0.36 & 0.80 & 0.35 & 0.87 \\
    \multirow{2}{*}{Grid-Difficult} & Alg.~\ref{algo_1} (Ours)
    & \textbf{0} & \textbf{0.76} & \textbf{1.00} & \textbf{0.30} & 0.93 \\
    & MonoMPC~\cite{11283048} & 70 & 0.31 & 0.74 & 0.40 & \textbf{0.91} \\
    \bottomrule
\end{tabular}
\end{table*}

\noindent \textbf{Comparing the Planning Performance:} 
Table~\ref{table_monompc} compares our planner with MonoMPC in Corridor,
Grid-Easy, and Grid-Difficult environments (Fig.~\ref{fig:real_world_snapshots}(a)-(c)),
showing how our improved collision model along with the learned planner
improves the overall navigation performance as compared to MonoMPC.
Our planner's observed collision rates are $7\%$, $3\%$, and $0\%$,
versus $37\%$, $23\%$, and $70\%$ for MonoMPC, respectively.
It also achieves more than twice the average linear speed
($0.76$--$0.81$\,m/s versus $0.28$--$0.36$\,m/s), while using
21 candidates per iteration and five CEM refinements, compared with
MonoMPC's 100 candidates and 20 refinements.

\begin{figure}[!t]
    \centering
    \includegraphics[width=0.95\columnwidth]
    {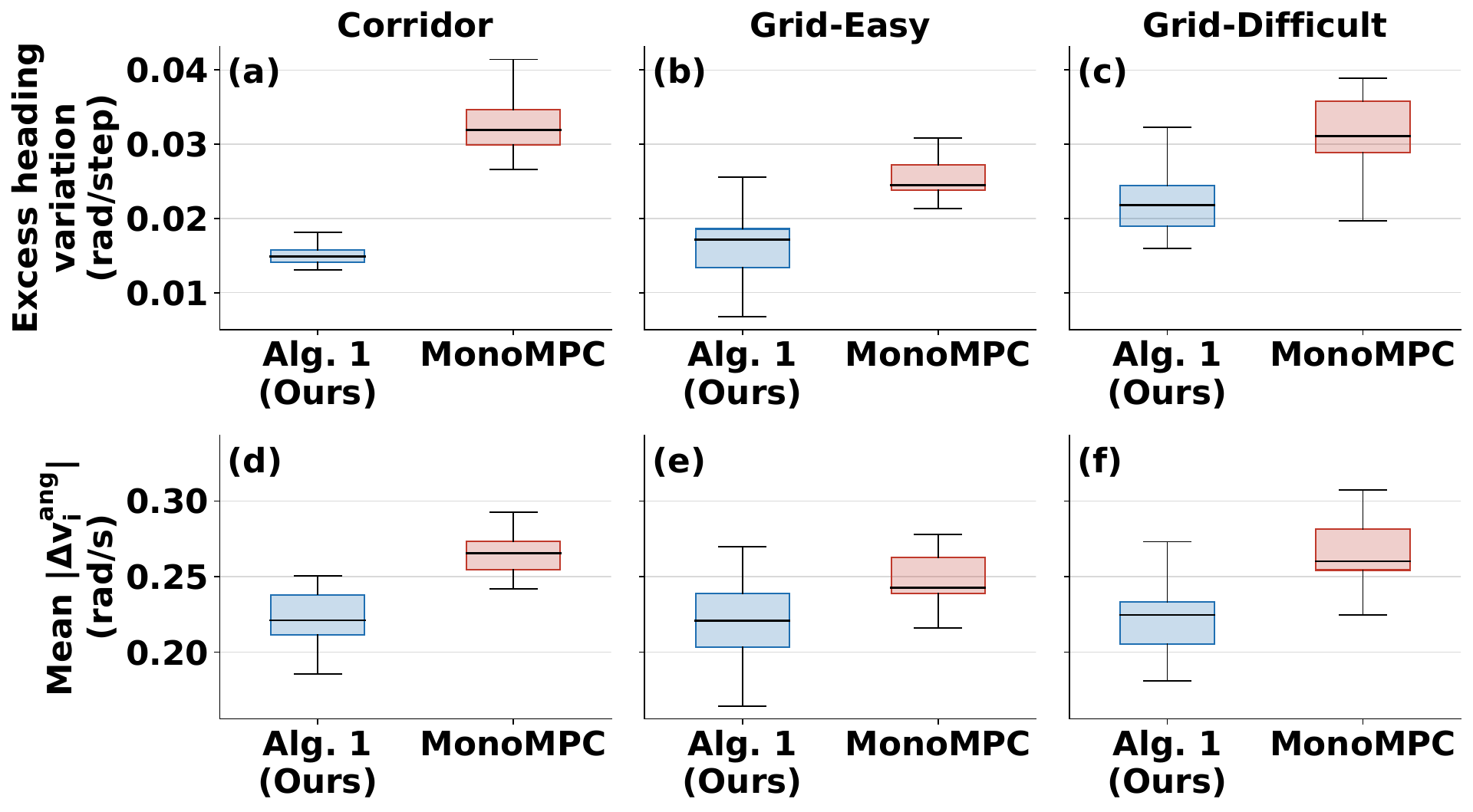}
    \caption{Per-run steering-oscillation metrics for Alg.~\ref{algo_1} 
    and MonoMPC, reported separately for the Corridor, Grid-Easy, and
    Grid-Difficult environments.
    The top row shows excess heading variation, while the bottom row
    shows mean absolute angular-velocity command variation.
    The lower distributions for our method indicate fewer back-and-forth
    heading fluctuations and less abrupt corrective steering input.}
    \label{fig:heading_omega}
    \Description{Box plots compare per-run excess heading variation and
    angular-command variation for the proposed planner 
    and MonoMPC across Corridor, Grid-Easy, and Grid-Difficult.}
    \vspace{-0.3cm}
\end{figure}

Figure~\ref{fig:heading_omega} compares excess heading variation and the mean absolute change in angular-velocity commands across individual runs. Our planner has lower medians for both metrics in all three environments, which indicates less heading oscillation and smaller steering corrections. This behavior is consistent with the use of observation history in our collision model. In cluttered environments, MonoMPC can overlook nearby obstacles outside the current view because its collision model uses only the latest image. As the robot turns, these obstacles can enter the camera's view and trigger further changes in direction. Our model uses earlier observations to account for previously seen obstacles, which can reduce the need for these repeated corrections.

\begin{figure}[!t]
    \centering
    \includegraphics[width=\columnwidth]{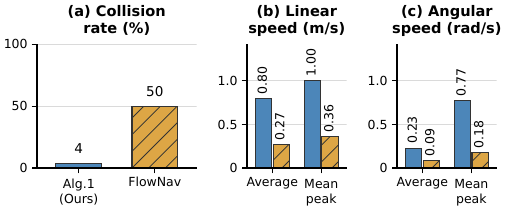}
    \caption{Comparison of Alg.~\ref{algo_1} (Ours) with FlowNav~\cite{gode2025flownav}
    in the Corridor environment (30 runs per method).}
    \label{fig:flownav}
    \Description{Three bar-plot panels compare Alg. 1 (Ours) with FlowNav
    in the Corridor environment. Panel (a) shows collision rates of 4 and
    50 percent, respectively. Panel (b) shows average and mean per-run peak
    linear speeds in metres per second; panel (c) shows the corresponding
    angular-speed statistics in radians per second.}
\end{figure}

\subsection{Comparison with FlowNav \texorpdfstring{\cite{gode2025flownav}}{}}
\label{sec:flownav}
\noindent FlowNav~\cite{gode2025flownav} is an end-to-end learned
visual-navigation policy. Building on the topological, goal-conditioned
architecture of NoMaD~\cite{10610665}, it replaces the diffusion-based
action head with conditional flow matching for faster action generation
and augments the visual observation tokens with depth priors obtained
from an off-the-shelf monocular depth foundation
model~\cite{NEURIPS2024_26cfdcd8}. Like other end-to-end policies,
FlowNav maps observations directly to trajectories and has no explicit
mechanism for evaluating the task risk of a candidate control. Therefore,
safety is only implicit in the imitation objective.
Fig.~\ref{fig:flownav} shows that the proposed local navigation policy
(Alg.~\ref{algo_1})
achieves higher average and mean per-run peak speeds than FlowNav
($0.80$ and $1.00$\,m/s versus $0.27$ and $0.36$\,m/s), while reducing
the observed collision rate from $50\%$ to $4\%$, a relative reduction
of $92\%$.
Thus, in these trials, the higher operating speeds are accompanied by
a lower observed collision rate. Refer to the videos in the supplementary 
material to further understand how the lack of explicit safety guidance during inference harms the performance of FlowNav. 

\begin{figure}[!htbp]
    \centering
    \includegraphics[width=\columnwidth]{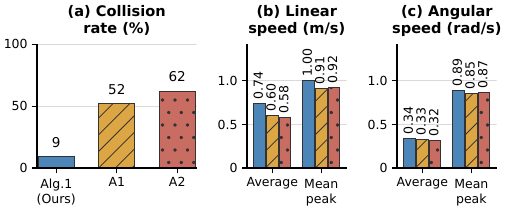}
    \caption{Ablation of Alg.~\ref{algo_1} (30 runs per method). 
    \emph{A1} and \emph{A2} use
    fixed initialization with learned and perturbed weights and temperature,
    respectively.}
    \label{fig:ablation}
    \Description{Three bar-plot panels compare Alg. 1 (Ours), A1 with fixed
    initialization and learned weights and temperature, and A2 with fixed
    initialization and perturbed weights and temperature. Panel (a) shows
    collision rates of 9, 52, and 62 percent, respectively. Panel (b) shows
    average and mean per-run peak linear speeds in metres per second;
    panel (c) shows the corresponding angular-speed statistics in radians
    per second.}
    \vspace{-0.6cm}
\end{figure}

\subsection{Ablation}
\label{sec:ablation}
\noindent We assess the contribution of learned initialization, cost weights, and 
temperature in Alg.~\ref{algo_1} using A1 and A2 (Fig.~\ref{fig:ablation}), 
both initialized with $\mathbf{u}_{\mathrm{mean}}^{(0)}=\mathbf{0}$ and 
fixed variance $\boldsymbol{\rho}^{(0)}$. The ablation uses a grid environment
(Fig.~\ref{fig:real_world_snapshots}(d)), more challenging than Grid-Difficult
(Fig.~\ref{fig:real_world_snapshots}(a)).

\noindent\textbf{A1:}
The first baseline retains the cost weights $\boldsymbol{w}_h,w_n$ and
temperature $\eta$ predicted by the policy network 
$\pi_{\boldsymbol{\theta}}$. 

\noindent\textbf{A2:}
The second baseline adds uniform
random noise to the policy-predicted cost weights $\boldsymbol{w}_h,w_n$
and temperature $\eta$ before running Alg.~\ref{algo_1}. As can be seen, the learned planner performs the best in terms of collision rate, speed and steering smoothness. The results are not surprising. Since, both optimizer warm-start and weights are learned together, even just removing the former (Ablation A1) drastically reduces the performance. Removing warm-start benefit and perturbing the learned weights (Ablation A2) leads to further degradation.

\subsection{Zero-Shot Transfer}
\label{sec:zero_shot}
\noindent Fig.~\ref{fig:real_world_snapshots}(e) shows zero-shot transfer 
of the unmodified trained collision model and policy
network to a Parrot Bebop drone (see supplementary video), and Alg.~\ref{algo_1} is used without
modification because the platform accepts the same velocity-command
parameterization.

\section{Conclusions and Future Work}

We presented a model-based imitation learning approach for monocular 
navigation in cluttered environments, integrating a learned collision model 
into a scene- and context-adaptive trajectory optimizer trained via imitation loss. 
The collision model predicts a worst-case clearance distribution from image history, 
helping account for previously observed, out-of-view obstacles without occupancy mapping.

Real-world tests showed lower collision rates and higher average speeds than 
MonoMPC and FlowNav, with smoother actions and fewer candidates per CEM iteration 
than MonoMPC. A drone demonstration qualitatively supported transfer of the full 
pipeline to a new robot and camera without modification or retraining.

Our evaluation is limited to static environments. Future work will extend 
the collision model to 
dynamic environments and test robustness to sensor occlusion 
and high-speed motion blur.

\noindent \textbf{AI Usage Acknowledgment:} We used GPT6Pro 
for text polishing
and formatting, figure editing, and code generation. All ideas and 
algorithmic development, experimental results
are our own.

\bibliographystyle{IEEEtran}
\bibliography{references}

@inproceedings{yin2023risk,
  title={Risk-aware model predictive path integral control using conditional value-at-risk},
  author={Yin, Ji and Zhang, Zhiyuan and Tsiotras, Panagiotis},
  booktitle={2023 IEEE International Conference on Robotics and Automation (ICRA)},
  pages={7937--7943},
  year={2023},
  organization={IEEE}
}

@inproceedings{jacquet2024n,
  title={{N-MPC} for Deep Neural Network-Based Collision Avoidance exploiting Depth Images},
  author={Jacquet, Martin and Alexis, Kostas},
  booktitle={2024 IEEE International Conference on Robotics and Automation (ICRA)},
  pages={13536--13542},
  year={2024},
  organization={IEEE}
}

@article{simon2018kernel,
  title={Kernel Distribution Embeddings: Universal Kernels,
         Characteristic Kernels and Kernel Metrics on Distributions},
  author={Simon-Gabriel, Carl-Johann and Sch{\"o}lkopf, Bernhard},
  journal={Journal of Machine Learning Research},
  volume={19},
  number={44},
  pages={1--29},
  year={2018}
}

@article{pagnoncelli2009sample,
  title={Sample average approximation method for chance constrained programming: theory and applications},
  author={Pagnoncelli, Bernardo K and Ahmed, Shabbir and Shapiro, Alexander},
  journal={Journal of Optimization Theory and Applications},
  volume={142},
  number={2},
  pages={399--416},
  year={2009},
  publisher={Springer}
}

@incollection{botev2013cross,
  title={The cross-entropy method for optimization},
  author={Botev, Zdravko I and Kroese, Dirk P and Rubinstein, Reuven Y and L’Ecuyer, Pierre},
  booktitle={Handbook of Statistics},
  volume={31},
  pages={35--59},
  year={2013},
  publisher={Elsevier}
}

@article{DBLP:journals/ral/LewBP24,
  author       = {Thomas Lew and
                  Riccardo Bonalli and
                  Marco Pavone},
  title        = {Risk-Averse Trajectory Optimization via Sample Average Approximation},
  journal      = {IEEE Robotics and Automation Letters},
  volume       = {9},
  number       = {2},
  pages        = {1500--1507},
  year         = {2024},
  url          = {https://doi.org/10.1109/LRA.2023.3331889},
  doi          = {10.1109/LRA.2023.3331889},
  bibsource    = {dblp computer science bibliography, https://dblp.org}
}

@book{shapiro2014lectures,
  title={Lectures on Stochastic Programming: Modeling and Theory},
  author={Shapiro, Alexander and Dentcheva, Darinka and Ruszczy{\'n}ski, Andrzej},
  edition={2},
  isbn={9781611973426},
  lccn={2014005486},
  series={MOS-SIAM Series on Optimization},
  doi={10.1137/1.9781611973433},
  year={2014},
  publisher={Society for Industrial and Applied Mathematics}
}

@article{10.5555/2188385.2188410,
  author = {Gretton, Arthur and Borgwardt, Karsten M. and Rasch, Malte J. and Sch\"{o}lkopf, Bernhard and Smola, Alexander},
  title = {A Kernel Two-Sample Test},
  journal = {Journal of Machine Learning Research},
  volume = {13},
  number = {25},
  pages = {723--773},
  year = {2012}
}

@misc{jax2018github,
  author = {James Bradbury and Roy Frostig and Peter Hawkins and Matthew James Johnson and Yash Katariya and Chris Leary and Dougal Maclaurin and George Necula and Adam Paszke and Jake Vander{P}las and Skye Wanderman-{M}ilne and Qiao Zhang},
  title = {{JAX}: composable transformations of {P}ython+{N}um{P}y programs},
  url = {https://github.com/jax-ml/jax},
  version = {0.3.13},
  year = {2018},
}

@inproceedings{Shirai_2023,
   title={Covariance Steering for Uncertain Contact-rich Systems},
   url={http://dx.doi.org/10.1109/ICRA48891.2023.10160249},
   DOI={10.1109/icra48891.2023.10160249},
   booktitle={2023 IEEE International Conference on Robotics and Automation (ICRA)},
   publisher={IEEE},
   author={Shirai, Yuki and Jha, Devesh K. and Raghunathan, Arvind U.},
   year={2023},
   pages={7923--7929},
   month=may }

@ARTICLE{10979910,
  author={Sharma, Basant and Singh, Arun Kumar},
  journal={IEEE Robotics and Automation Letters}, 
  title={Trajectory Optimization Under Stochastic Dynamics Leveraging Maximum Mean Discrepancy}, 
  year={2025},
  volume={10},
  number={6},
  pages={6079-6086},
  doi={10.1109/LRA.2025.3565335}}

@inproceedings{depthanything,
  title={{Depth Anything}: Unleashing the Power of Large-Scale Unlabeled Data},
  author={Yang, Lihe and Kang, Bingyi and Huang, Zilong and Xu, Xiaogang and Feng, Jiashi and Zhao, Hengshuang},
  booktitle={Proceedings of the IEEE/CVF Conference on Computer Vision and Pattern Recognition (CVPR)},
  year={2024},
  pages={10371--10381}
}

@misc{https://doi.org/10.48550/arxiv.2302.12288,
  doi = {10.48550/ARXIV.2302.12288},
  url = {https://arxiv.org/abs/2302.12288},
  author = {Bhat, Shariq Farooq and Birkl, Reiner and Wofk, Diana and Wonka, Peter and Müller, Matthias},
  title = {{ZoeDepth}: Zero-shot Transfer by Combining Relative and Metric Depth},
  howpublished = {arXiv preprint arXiv:2302.12288},
  year = {2023},
  copyright = {arXiv.org perpetual, non-exclusive license}
}

@incollection{simon2023mononav,
  author = {Simon, Nathaniel and Majumdar, Anirudha},
  title = {{MonoNav}: {MAV} Navigation via Monocular Depth Estimation and Reconstruction},
  booktitle = {Experimental Robotics: The 18th International Symposium},
  series = {Springer Proceedings in Advanced Robotics},
  volume = {30},
  pages = {415--426},
  publisher = {Springer},
  year = {2024},
  doi = {10.1007/978-3-031-63596-0_37}
}

@INPROCEEDINGS{10610665,
  author={Sridhar, Ajay and Shah, Dhruv and Glossop, Catherine and Levine, Sergey},
  booktitle={2024 IEEE International Conference on Robotics and Automation (ICRA)}, 
  title={{NoMaD}: Goal Masked Diffusion Policies for Navigation and Exploration},
  year={2024},
  volume={},
  number={},
  pages={63-70},
  doi={10.1109/ICRA57147.2024.10610665}}

@inproceedings{chaplot2020learning,
  title={Learning To Explore Using Active Neural SLAM},
  author={Chaplot, Devendra Singh and Gandhi, Dhiraj and Gupta, Saurabh and Gupta, Abhinav and Salakhutdinov, Ruslan},
  booktitle={International Conference on Learning Representations (ICLR)},
  year={2020}
}

@ARTICLE{11084873,
  author={Sharma, Basant and Singh, Arun Kumar},
  journal={IEEE Transactions on Automation Science and Engineering}, 
  title={{MMD-OPT}: Maximum Mean Discrepancy Based Sample Efficient Collision Risk Minimization for Autonomous Driving},
  year={2025},
  volume={22},
  number={},
  pages={19051--19068},
  doi={10.1109/TASE.2025.3589941}}

@inproceedings{NEURIPS2024_26cfdcd8,
 author = {Yang, Lihe and Kang, Bingyi and Huang, Zilong and Zhao, Zhen and Xu, Xiaogang and Feng, Jiashi and Zhao, Hengshuang},
 booktitle = {Advances in Neural Information Processing Systems},
 editor = {A. Globerson and L. Mackey and D. Belgrave and A. Fan and U. Paquet and J. Tomczak and C. Zhang},
 pages = {21875--21911},
 publisher = {Curran Associates, Inc.},
 title = {{Depth Anything V2}},
 volume = {37},
 year = {2024}
}

@misc{kidger2021equinox,
    author={Patrick Kidger and Cristian Garcia},
    title={{E}quinox: neural networks in {JAX} via callable {P}y{T}rees and filtered transformations},
    year={2021},
    howpublished={\emph{Differentiable Programming workshop at Neural Information Processing Systems 2021}}
}

@InProceedings{pmlr-v119-amos20a,
  title = 	 {The Differentiable Cross-Entropy Method},
  author =       {Amos, Brandon and Yarats, Denis},
  booktitle = 	 {Proceedings of the 37th International Conference on Machine Learning},
  pages = 	 {291--302},
  year = 	 {2020},
  editor = 	 {III, Hal Daumé and Singh, Aarti},
  volume = 	 {119},
  series = 	 {Proceedings of Machine Learning Research},
  month = 	 {13--18 Jul},
  publisher =    {PMLR}
}

@INPROCEEDINGS{gode2025flownav,
  author={Gode, Samiran and Nayak, Abhijeet and Oliveira, Débora N.P. and Krawez, Michael and Schmid, Cordelia and Burgard, Wolfram},
  booktitle={2025 IEEE/RSJ International Conference on Intelligent Robots and Systems (IROS)}, 
  title={{FlowNav}: Combining Flow Matching and Depth Priors for Efficient Navigation},
  year={2025},
  pages={17762--17768},
  doi={10.1109/IROS60139.2025.11246634}
}

@ARTICLE{11283048,
  author={Sharma, Basant and Jadhav, Prajyot and Paul, Pranjal and Krishna, K.Madhava and Singh, Arun Kumar},
  journal={IEEE Robotics and Automation Letters}, 
  title={{MonoMPC}: Monocular Vision Based Navigation With Learned Collision Model and Risk-Aware Model Predictive Control},
  year={2026},
  volume={11},
  number={2},
  pages={1330-1337},
  doi={10.1109/LRA.2025.3641112}}

@InProceedings{kim2025care,
  title = {{CARE}: Enhancing Safety of Visual Navigation through Collision Avoidance via Repulsive Estimation},
  author = {Kim, Joonkyung and Sim, Joonyeol and Kim, Woojun and Sycara, Katia P. and Nam, Changjoo},
  booktitle = {Proceedings of The 9th Conference on Robot Learning},
  pages = {3704--3719},
  year = {2025},
  editor = {Lim, Joseph and Song, Shuran and Park, Hae-Won},
  volume = {305},
  series = {Proceedings of Machine Learning Research},
  publisher = {PMLR}
}

@inproceedings{agrawal2017discrete,
  title={Discrete Control Barrier Functions for Safety-Critical Control of Discrete Systems with Application to Bipedal Robot Navigation},
  author={Agrawal, Ayush and Sreenath, Koushil},
  booktitle={Proceedings of Robotics: Science and Systems},
  year={2017},
  address={Cambridge, Massachusetts},
  month=jul,
  doi={10.15607/RSS.2017.XIII.073}
}

@INPROCEEDINGS{9483029,
  author={Zeng, Jun and Zhang, Bike and Sreenath, Koushil},
  booktitle={2021 American Control Conference (ACC)}, 
  title={Safety-Critical Model Predictive Control with Discrete-Time Control Barrier Function}, 
  year={2021},
  volume={},
  number={},
  pages={3882-3889},
  doi={10.23919/ACC50511.2021.9483029}}

@inproceedings{qi2017pointnet,
  title={{PointNet}: Deep Learning on Point Sets for {3D} Classification and Segmentation},
  author={Qi, Charles R and Su, Hao and Mo, Kaichun and Guibas, Leonidas J},
  booktitle={Proceedings of the IEEE Conference on Computer Vision and Pattern Recognition (CVPR)},
  pages={652--660},
  year={2017}
}

@article{Andrejev_2025,
   title={$\pi$-MPPI: A Projection-Based Model Predictive Path Integral Scheme for Smooth Optimal Control of Fixed-Wing Aerial Vehicles},
   volume={10},
   ISSN={2377-3774},
   url={http://dx.doi.org/10.1109/LRA.2025.3568323},
   DOI={10.1109/lra.2025.3568323},
   number={6},
   journal={IEEE Robotics and Automation Letters},
   publisher={Institute of Electrical and Electronics Engineers (IEEE)},
   author={Andrejev, Edvin Martin and Manoharan, Amith and Unt, Karl-Eerik and Singh, Arun Kumar},
   year={2025},
   pages={6496--6503}}

@inproceedings{gupta2017cognitive,
  title={Cognitive Mapping and Planning for Visual Navigation},
  author={Gupta, Saurabh and Davidson, James and Levine, Sergey and Sukthankar, Rahul and Malik, Jitendra},
  booktitle={Proceedings of the IEEE Conference on Computer Vision and Pattern Recognition (CVPR)},
  pages={2616--2625},
  year={2017}
}

@article{cartillier2021semantic,
  author  = {Cartillier, Vincent and Ren, Zhile and Jain, Neha and Lee, Stefan
             and Essa, Irfan and Batra, Dhruv},
  title   = {Semantic {MapNet}: Building Allocentric Semantic Maps and
             Representations from Egocentric Views},
  journal = {Proceedings of the AAAI Conference on Artificial Intelligence},
  volume  = {35},
  number  = {2},
  pages   = {964--972},
  year    = {2021},
  doi     = {10.1609/aaai.v35i2.16180}
}

@inproceedings{kahn2018self,
  author    = {Kahn, Gregory and Villaflor, Adam and Ding, Bosen and
               Abbeel, Pieter and Levine, Sergey},
  title     = {Self-Supervised Deep Reinforcement Learning with Generalized
               Computation Graphs for Robot Navigation},
  booktitle = {2018 IEEE International Conference on Robotics and Automation (ICRA)},
  year      = {2018},
  pages     = {5129--5136},
  publisher = {IEEE},
  doi       = {10.1109/ICRA.2018.8460655},
  url       = {https://arxiv.org/abs/1709.10489}
}

@inproceedings{pan2020zero,
  author    = {Pan, Xinlei and Zhang, Tingnan and Ichter, Brian and
               Faust, Aleksandra and Tan, Jie and Ha, Sehoon},
  title     = {Zero-Shot Imitation Learning from Demonstrations for
               Legged Robot Visual Navigation},
  booktitle = {2020 IEEE International Conference on Robotics and Automation (ICRA)},
  year      = {2020},
  pages     = {679--685},
  publisher = {IEEE},
  doi       = {10.1109/ICRA40945.2020.9196602},
  url       = {https://arxiv.org/abs/1909.12971}
}

@inproceedings{acerbo2024driving,
  author    = {Acerbo, Flavia Sofia and Swevers, Jan and Tuytelaars, Tinne
               and Son, Tong Duy},
  title     = {Driving from Vision through Differentiable Optimal Control},
  booktitle = {2024 IEEE/RSJ International Conference on Intelligent Robots and Systems (IROS)},
  year      = {2024},
  pages     = {2824--2831},
  publisher = {IEEE},
  doi       = {10.1109/IROS58592.2024.10802306},
  url       = {https://arxiv.org/abs/2403.15102}
}

@inproceedings{pmlr-v80-srinivas18b,
  author    = {Srinivas, Aravind and Jabri, Allan and Abbeel, Pieter and
               Levine, Sergey and Finn, Chelsea},
  title     = {Universal Planning Networks: Learning Generalizable
               Representations for Visuomotor Control},
  booktitle = {Proceedings of the 35th International Conference on
               Machine Learning},
  series    = {Proceedings of Machine Learning Research},
  volume    = {80},
  pages     = {4732--4741},
  year      = {2018},
  publisher = {PMLR},
  url       = {https://proceedings.mlr.press/v80/srinivas18b.html}
}

@inproceedings{amos2018differentiablempc,
  author    = {Amos, Brandon and Jimenez, Ivan and Sacks, Jacob and
               Boots, Byron and Kolter, J. Zico},
  title     = {Differentiable {MPC} for End-to-end Planning and Control},
  booktitle = {Advances in Neural Information Processing Systems},
  volume    = {31},
  pages     = {8299--8310},
  year      = {2018},
  url       = {https://papers.nips.cc/paper/2018/hash/ba6d843eb4251a4526ce65d1807a9309-Abstract.html}
}

@inproceedings{hu2022mile,
  author    = {Hu, Anthony and Corrado, Gianluca and Griffiths, Nicolas and
               Murez, Zachary and Gurau, Corina and Yeo, Hudson and
               Kendall, Alex and Cipolla, Roberto and Shotton, Jamie},
  title     = {Model-Based Imitation Learning for Urban Driving},
  booktitle = {Advances in Neural Information Processing Systems},
  volume    = {35},
  pages     = {20703--20716},
  year      = {2022},
  doi       = {10.52202/068431-1505}
}

@article{huang2024dipp,
  author  = {Huang, Zhiyu and Liu, Haochen and Wu, Jingda and Lv, Chen},
  title   = {Differentiable Integrated Motion Prediction and Planning
             With Learnable Cost Function for Autonomous Driving},
  journal = {IEEE Transactions on Neural Networks and Learning Systems},
  volume  = {35},
  number  = {11},
  pages   = {15222--15236},
  year    = {2024},
  doi     = {10.1109/TNNLS.2023.3283542},
  url     = {https://arxiv.org/abs/2207.10422}
}

\clearpage
\twocolumn[{
    \centering
    \LARGE\bfseries Supplementary Material\par
    \vspace{1.5em}
}]
\pdfbookmark[0]{Supplementary Material}{supplementary-material}

\appendices
\setcounter{equation}{0}
\setcounter{figure}{0}
\setcounter{table}{0}
\renewcommand{\theequation}{S\arabic{equation}}
\renewcommand{\thefigure}{S\arabic{figure}}
\renewcommand{\thetable}{S\arabic{table}}
\renewcommand{\theHequation}{supp.\arabic{equation}}
\renewcommand{\theHfigure}{supp.\arabic{figure}}
\renewcommand{\theHtable}{supp.\arabic{table}}

\section{Introduction}
\label{sec:supp_introduction}

This supplementary material uses the notation of the main paper.
Section~\ref{appendix} provides the collision-risk formulation and analysis
supporting main-paper Section~\ref{sec:training}.
Section~\ref{sec:network_implementation} specifies the implemented network
architectures for main-paper Sections~\ref{sec:collision_architecture}
and~\ref{sec:planning_algorithm}.
Section~\ref{sec:qualitative_comparisons} presents qualitative results
corresponding to the comparisons, ablations, and transfer experiments in
main-paper Section~\ref{sec:validation}.
It also includes a quantitative comparison with CARE supporting the
discussion in main-paper Section~\ref{sec:introduction}.
Section~\ref{sec:evaluation_metrics} defines the evaluation metrics and
their aggregation across runs.
Table~\ref{tab:supp_main_mapping} summarizes these connections.

\begin{table}[H]
    \centering
    \caption{Guide to the supplementary material and corresponding
    main-paper content.}
    \label{tab:supp_main_mapping}
    \begin{tabularx}{\columnwidth}{@{}>{\raggedright\arraybackslash}p{0.54\columnwidth}>{\raggedright\arraybackslash}X@{}}
        \toprule
        Supplementary material & Main paper \\
        \midrule
        Sec.~\ref{appendix}: Collision-risk formulation and analysis
        & Sec.~\ref{sec:training};
          Eqs.~\eqref{clearance_distributions}--\eqref{barrier_neural_cost} \\
        Secs.~\ref{sec:network_inputs}--\ref{sec:policy_implementation}:
        Network architecture
        & Secs.~\ref{sec:collision_architecture},
          \ref{sec:planning_algorithm};
          Figs.~\ref{fig:collision_model_pipeline},
          \ref{fig:diffcem_pipeline} \\
        Secs.~\ref{sec:training_configuration}--\ref{sec:computation_time}:
        Training and computation time
        & Secs.~\ref{sec:planning_algorithm},
          \ref{sec:implementation}; Alg.~\ref{algo_1} \\
        Sec.~\ref{sec:supp_monompc}: MonoMPC trajectories and oscillations
        & Sec.~\ref{sec:monompc}; Table~\ref{table_monompc};
          Fig.~\ref{fig:heading_omega} \\
        Sec.~\ref{sec:supp_flownav}: FlowNav trajectories
        & Sec.~\ref{sec:flownav}; Fig.~\ref{fig:flownav} \\
        Sec.~\ref{sec:supp_care}: CARE comparison and trajectories
        & Sec.~\ref{sec:introduction}: Safety augmentation of visual policies \\
        Sec.~\ref{sec:supp_fixed_init}: Fixed-initialization baselines
        & Sec.~\ref{sec:ablation}; Fig.~\ref{fig:ablation} \\
        Sec.~\ref{sec:supp_bebop}: Bebop transfer
        & Sec.~\ref{sec:zero_shot}; Fig.~\ref{fig:real_world_snapshots}(e) \\
        Sec.~\ref{sec:evaluation_metrics}: Evaluation metrics
        & Sec.~\ref{sec:validation}; Table~\ref{table_monompc};
          Figs.~\ref{fig:heading_omega}--\ref{fig:ablation} \\
        \bottomrule
    \end{tabularx}
\end{table}

\section{Collision-Risk Formulation and Analysis}\label{appendix}

\subsection{Deterministic Gaussian Quantiles}
\label{sec:clearance_quantiles}

Evaluating $r^{\mathrm{emp}}$ and the neural collision-risk cost in
\eqref{barrier_neural_cost} requires a finite set of residual quantiles
$\{{}^{j}\overline h_b(
\mathbf{o},\mathbf{u};\boldsymbol{\psi},\boldsymbol{\alpha})\}_{j=1}^{N}$.
We use deterministic Gaussian quantiles instead of Monte Carlo samples to
avoid sampling variability. For fixed inputs, the quantile locations and
residual quantiles are unchanged, so repeated evaluations produce the same
value of $r^{\mathrm{emp}}$ without stochasticity from Monte Carlo sampling.

We use $N$ fixed midpoint locations of the standard-normal distribution:
\begin{align}
    {}^{j}z
    &=\Phi^{-1}\!\left(\frac{j-0.5}{N}\right),
    \qquad j=1,\ldots,N,
    \label{normal_quantiles}
\end{align}
where $\Phi^{-1}$ is the inverse standard-normal cumulative distribution
function. Using the fixed locations in \eqref{normal_quantiles}, the $h_b$ and
$\overline h_b$ quantiles are
\begin{subequations}
\label{barrier_quantiles}
\begin{align}
    {}^{j}h_b(\mathbf{o},\mathbf{u};\boldsymbol{\psi},\boldsymbol{\alpha})
    &=\mu_{h_b}+\sigma_{h_b}\,{}^{j}z,
    \label{barrier_value_quantiles}\\
    {}^{j}\overline h_b(\mathbf{o},\mathbf{u};\boldsymbol{\psi},\boldsymbol{\alpha})
    &=\max\!\left(0,{}^{j}h_b\right),
    \qquad j=1,\ldots,N,
    \label{barrier_residual_quantiles}
\end{align}
\end{subequations}
where the mean and variance of
$h_b(\mathbf{o},\mathbf{u};\boldsymbol{\psi},\boldsymbol{\alpha})$ are
\begin{subequations}
\label{barrier_distribution}
\begin{align}
    \mu_{h_b}(\mathbf{o},\mathbf{u};\boldsymbol{\psi},\boldsymbol{\alpha})
    &=-\mu_{\boldsymbol{\psi}}(\mathbf{o},\mathbf{u})
      +d_{\mathrm{thresh}}
      \notag\\*
    &\quad -(1-\gamma)\left(
        -\mu_{0,\boldsymbol{\alpha}}(\mathbf{o})
        +d_{\mathrm{thresh}}
      \right),
    \label{barrier_mean}\\
    \sigma_{h_b}^{2}(\mathbf{o},\mathbf{u};\boldsymbol{\psi},\boldsymbol{\alpha})
    &=\sigma_{\boldsymbol{\psi}}^{2}(\mathbf{o},\mathbf{u})
      +(1-\gamma)^2
      \sigma_{0,\boldsymbol{\alpha}}^{2}(\mathbf{o}).
    \label{barrier_variance}
\end{align}
\end{subequations}
The models parameterized by $\boldsymbol{\psi}$ and $\boldsymbol{\alpha}$
separately predict the marginal distributions of
$d(\mathbf{o},\mathbf{u};\boldsymbol{\psi})$ and
$d_0(\mathbf{o};\boldsymbol{\alpha})$ in
\eqref{clearance_distributions}. Because these models do not define a joint
distribution over the two clearances, we treat their predictions as
independent. This assumption yields the moments in
\eqref{barrier_distribution}.

\subsection{MMD-Based Collision-Risk Surrogate}
\label{sec:mmd_surrogate}

Let $p_{\overline h_b}$ denote the distribution of the nonnegative barrier
residual $\overline h_b=[h_b]_+$ represented by the quantiles in
\eqref{barrier_residual_quantiles}. The key observation is that its entire probability mass
lies on the nonnegative side of
$\overline h_b=0$. Moreover, as the collision risk
$\mathbb{P}(h_b>0)$ approaches zero, $p_{\overline h_b}$ converges to the
Dirac measure $p_{\delta_0}$ concentrated at the origin
\cite{11084873}. Thus, collision risk can be reduced by minimizing the
difference between $p_{\overline h_b}$ and $p_{\delta_0}$. We formulate this distribution matching
through MMD~\cite{10.5555/2188385.2188410,simon2018kernel} and use the MMD-based
collision-risk surrogate proposed in
\cite{11283048,11084873,10979910}. For the
distribution $p_{\overline h_b}$, let
$\mu[p_{\overline h_b}]=\mathbb{E}_{z\sim p_{\overline h_b}}[K_{\lambda}(z,\cdot)]$ denote its reproducing
kernel Hilbert space (RKHS) embedding. The resulting collision-risk surrogate is
\begin{align}
    r_{\mathrm{MMD}}
    &\triangleq
    \operatorname{MMD}^{2}\!\left(p_{\overline h_b},p_{\delta_0}\right)
    \notag\\
    &=
    \left\|
        \mu[p_{\overline h_b}]-\mu[p_{\delta_0}]
    \right\|_{\mathcal H}^{2}.
    \label{risk_mmd}
\end{align}
The Laplace kernel used in this work is
\begin{align}
    K_{\lambda}(z,z')
    &=\exp\!\left(-\frac{|z-z'|}{\lambda}\right).
    \label{mmd_kernel}
\end{align}
The Laplace kernel is characteristic. Therefore,
$r_{\mathrm{MMD}}=0$ if and only if
$p_{\overline h_b}=p_{\delta_0}$~\cite{10.5555/2188385.2188410}, which is
equivalent to $\mathbb{P}(h_b>0)=0$, or equivalently, zero collision risk
in \eqref{risk_def}.


For efficient evaluation, we replace the population embeddings in
\eqref{risk_mmd} with a finite-quantile approximation based on the $N$
deterministic residual quantiles in \eqref{barrier_residual_quantiles}. This
converts the population expectations into finite sums and provides repeatable
surrogate values and gradients without Monte Carlo sampling variability. Let
\begin{align}
    \widehat P_N^{\overline h_b}
    &\triangleq \frac{1}{N}\sum_{j=1}^{N}
    \delta_{{}^{j}\overline h_b}
    \label{empirical_residual_measure}
\end{align}
denote the discrete residual measure that approximates
$p_{\overline h_b}$. The corresponding finite approximations of the kernel
mean embeddings are
\begin{equation}
\begin{aligned}
    \widehat\mu_N[\widehat P_N^{\overline h_b}]
    &=\frac{1}{N}\sum_{j=1}^{N}
      K_{\lambda}({}^{j}\overline h_b,\cdot),\\
    \widehat\mu_N[p_{\delta_0}]
    &=\frac{1}{N}\sum_{j=1}^{N}K_{\lambda}(0,\cdot).
\end{aligned}
\label{finite_kernel_embeddings}
\end{equation}
The finite-quantile approximation of $r_{\mathrm{MMD}}$ is
\begin{align}
    r_{\mathrm{MMD}}^{\mathrm{emp}}
    &=\left\|
        \widehat\mu_N[\widehat P_N^{\overline h_b}]
        -\widehat\mu_N[p_{\delta_0}]
      \right\|_{\mathcal H}^{2}.
    \label{empirical_mmd_definition}
\end{align}
Using the kernel trick, \eqref{empirical_mmd_definition} becomes
\begin{align}
    r_{\mathrm{MMD}}^{\mathrm{emp}}
    &=\frac{1}{N^2}\sum_{i=1}^{N}\sum_{j=1}^{N}
      K_{\lambda}({}^{i}\overline h_b,{}^{j}\overline h_b)
      \notag\\
    &\quad-\frac{2}{N}\sum_{i=1}^{N}
      K_{\lambda}({}^{i}\overline h_b,0)+1.
    \label{mmd_embedding}
\end{align}

\subsubsection{Finite-Quantile Convergence of \texorpdfstring{$r^{\mathrm{emp}}_{\mathrm{MMD}}$}{r emp MMD}}

Let
\[
    \widehat P_N^{h_b}\triangleq
    N^{-1}\sum_{j=1}^{N}\delta_{{}^{j}h_b}
\]
be the equal-weight measure supported
on the midpoint quantiles in \eqref{barrier_value_quantiles}. This midpoint
construction gives $\widehat P_N^{h_b}\longrightarrow p_{h_b}$ weakly, where
$p_{h_b}$ is the Gaussian barrier distribution specified by
\eqref{barrier_distribution}. Weak convergence is preserved under the
continuous transformation $h_b\mapsto\max(0,h_b)$ used in
\eqref{barrier_residual_quantiles}. Therefore,
$\widehat P_N^{\overline h_b}\longrightarrow p_{\overline h_b}$ weakly as
$N\longrightarrow\infty$.
For a fixed control sequence, the Laplace kernel is bounded and continuous.
Lemma~10 of \cite{simon2018kernel} therefore
implies
\begin{align}
    \left\|
      \widehat\mu_N[\widehat P_N^{\overline h_b}]
      -\mu[p_{\overline h_b}]
    \right\|_{\mathcal H}
    &\longrightarrow 0
    \quad\text{as}\quad N\longrightarrow\infty.
    \label{embedding_convergence}
\end{align}
Since $p_{\delta_0}$ is concentrated at zero, both its population embedding
and its $N$-point empirical embedding equal $K_{\lambda}(0,\cdot)$. Therefore,
$\widehat\mu_N[p_{\delta_0}]=\mu[p_{\delta_0}]$ for every $N$. Continuity of
the squared RKHS norm then gives
\begin{align}
    r_{\mathrm{MMD}}^{\mathrm{emp}}
    &\longrightarrow r_{\mathrm{MMD}}
    \quad\text{as}\quad N\longrightarrow\infty.
    \label{mmd_convergence}
\end{align}

\noindent\textbf{Importance of the kernel parameter.}
The kernel width $\lambda$ controls the sensitivity of $r_{\mathrm{MMD}}^{\mathrm{emp}}$. An
excessively large $\lambda$ makes the embeddings nearly constant and less
sensitive to collision risk. An excessively small $\lambda$ makes the
embeddings sharply separated and can produce overly conservative behavior.
We therefore predict a positive, control-dependent kernel width
$\lambda_{\boldsymbol{\psi}}(\mathbf{o},\mathbf{u})$ from the observation and
control sequence. Its output is bounded to
$\lambda_{\boldsymbol{\psi}}\in[0.05,25.0]$.

\noindent\textbf{Use across stages.}
During collision-model training,
$r_{\mathrm{MMD}}^{\mathrm{emp}}$ is evaluated on the residual quantiles
$\{{}^{j}\overline h\}_{j=1}^{N}$ computed directly from raw clearance:
${}^{j}\overline h=\max(0,d_{\mathrm{thresh}}-{}^{j}d)$, with
${}^{j}d=\mu_{\boldsymbol{\psi}}(\mathbf{o},\mathbf{u})
+\sigma_{\boldsymbol{\psi}}(\mathbf{o},\mathbf{u})\,{}^{j}z$ and
${}^{j}z$ given by \eqref{normal_quantiles}. This training stage does not
use the barrier formulation; see main-paper
Section~\ref{sec:collision_training}. Its gradients update
$\boldsymbol{\psi}$, including the predicted clearance uncertainty and
kernel width. During policy training, the collision models are frozen and
the same squared-MMD construction is evaluated on the barrier residual
quantiles $\{{}^{j}\overline h_b\}_{j=1}^{N}$ from
\eqref{barrier_residual_quantiles}. The resulting surrogate is used by
$J_{\mathrm{neural}}$ in \eqref{barrier_neural_cost}, while gradients update
only the policy parameters $\boldsymbol{\theta}$. At deployment, all learned
models are frozen and the surrogate is used only to evaluate control
sequences.

\section{Network Architecture and Implementation Details}
\label{sec:network_implementation}

The specifications below correspond to the collision model in main-paper
Fig.~\ref{fig:collision_model_pipeline} and the policy network
$\pi_{\boldsymbol{\theta}}$ in main-paper Fig.~\ref{fig:diffcem_pipeline}.
All tensor dimensions omit the batch axis. The implementation uses JAX and
Equinox with 32-bit floating-point parameters.

\subsection{Inputs and Normalization}
\label{sec:network_inputs}

\noindent\textbf{Visual input.}
The input contains $21$ point clouds: $m=20$ past frames followed by the
current frame, sampled at $0.1$\,s intervals. Each cloud has $300$ points
with two coordinates, giving an input tensor of size $21\times300\times2$.
Each point cloud is expressed in the robot frame at the time its image
was captured; the history is not transformed into a common frame.
Metric depth is obtained with the frozen Depth Anything V2 model and
back-projected as described in main-paper
Section~\ref{sec:implementation}. Point coordinates are normalized
independently by coordinate-wise training-set minima and maxima. The
collision model and policy use their respective saved normalization
statistics.

\noindent\textbf{State, goal, and control inputs.}
The state input contains the measured current linear and angular velocities
in $\mathbb{R}^2$; the policy additionally receives the local
goal $\mathbf{g}\in\mathbb{R}^2$. These vectors are standardized using
training-set means and standard deviations. The collision model receives
a flattened representation of the main-paper control sequence
$\mathbf{u}=[\mathbf{u}_0,\ldots,\mathbf{u}_{H-1}]
\in\mathbb{R}^{2\times H}$, with $H=50$ and
$\mathbf{u}_k=(v_k,\omega_k)^\top$. The flattened input is
$(v_0,\ldots,v_{49},\omega_0,\ldots,\omega_{49})\in\mathbb{R}^{100}$,
standardized using the corresponding linear- and angular-velocity
statistics. All $100$ control entries are optimized; the measured current
velocity is supplied separately through the state input. Velocity history
is not an input to the networks.

\subsection{Point-Cloud and Temporal Encoders}
\label{sec:perception_implementation}

The collision model and policy have independent encoders, with context
dimensions $D=40$ and $D=64$, respectively. Within each encoder, the same
point-cloud network processes all $21$ frames. Its layer specification is
given in Table~\ref{tab:supp_perception_layers}.

\begin{table}[!ht]
    \centering
    \caption{Perception encoder used independently by each network.
    $D=40$ for the collision model and $D=64$ for the policy.
    Arrows denote successive channel dimensions.}
    \label{tab:supp_perception_layers}
    \begin{tabularx}{\columnwidth}{@{}lX@{}}
        \toprule
        Component & Layer specification \\
        \midrule
        Point-wise encoder & Five Conv1D layers:
        $2\!\to\!64\!\to\!64\!\to\!64\!\to\!128\!\to\!D$ \\
        Pooling & Global maximum over the $300$ points \\
        Pooled projection & Linear $D\to256$, LayerNorm, ReLU,
        dropout, linear $256\to D$ \\
        Frame projection & Linear $D\to D$ \\
        Position embedding & Learned array of size $21\times D$ \\
        Temporal attention & One pre-normalized self-attention block;
        four heads of width $D/4$ \\
        Temporal MLP & Two linear layers, $D\to4D\to D$,
        with SiLU between them \\
        Context readout & Final (current-frame) token of dimension $D$ \\
        \bottomrule
    \end{tabularx}
\end{table}

\noindent\textbf{Point-cloud layers.}
All five convolutions use kernel size $1$, stride $1$, no padding,
and a bias, followed by ReLU. There is no per-point normalization.
LayerNorm is applied only to the pooled $256$-dimensional projection.
The pooled output and frame projections have biases and no output
activation. Dropout after the pooled ReLU has probability $0.1$.

\noindent\textbf{Temporal block.}
The learned position embeddings are added to the projected frame tokens.
Self-attention operates on the full $21$-token sequence without a causal
mask. Query, key, value, and output projections are $D\to D$ linear maps
without biases. Each head has dimension $10$ for the collision model and
$16$ for the policy. The attention and feed-forward sublayers each have
pre-LayerNorm and a residual connection. Dropout with probability $0.1$
is applied to the attention weights and to both residual-branch outputs.
The feed-forward hidden dimensions are $160$ and $256$, respectively.
All LayerNorm layers use learned affine parameters and
$\epsilon=10^{-5}$. The readout is
$\mathbf{c}_{\mathrm{feat}}\in\mathbb{R}^{40}$ for the collision model and
$\widetilde{\mathbf{c}}_{\mathrm{feat}}\in\mathbb{R}^{64}$ for the policy.

\subsection{Clearance and Kernel-Width Network}
\label{sec:clearance_implementation}

\noindent\textbf{Input and hidden layers.}
The concatenation of current velocity, candidate control sequence, and
visual context has dimension $2+100+40=142$. Four fully connected layers
have widths $142\to256\to256\to256\to40$. Each layer has a bias and is
followed by LayerNorm, SiLU, and dropout with probability $0.1$.

\noindent\textbf{Output heads.}
Three independent linear $40\to1$ heads with biases produce raw scalar
outputs $a_\mu$, $a_\sigma$, and $a_\lambda$. The output transformations are
\begin{align*}
    \mu_{\boldsymbol{\psi}} &= \operatorname{softplus}(a_\mu),\\
    \sigma_{\boldsymbol{\psi}} &= \operatorname{softplus}(a_\sigma)+10^{-3},\\
    \lambda_{\boldsymbol{\psi}} &= 0.05+24.95\operatorname{sigmoid}(a_\lambda).
\end{align*}
The standard-deviation output is squared wherever
$\sigma_{\boldsymbol{\psi}}^2$ is required. The kernel width lies in
$[0.05,25.0]$. The MMD risk is computed from the predicted distribution
using $N=20$ deterministic quantiles and has no additional trainable head.

\subsection{Policy and Soft-CEM Parameter Heads}
\label{sec:policy_implementation}

The policy concatenates its $64$-dimensional visual context with separate
$64$-dimensional state and goal embeddings. The resulting
$192$-dimensional vector is processed by a shared fusion MLP and five
output heads, as specified in Table~\ref{tab:supp_policy_layers}.

\begin{table}[!ht]
    \centering
    \caption{Policy layers following the independent perception encoder.
    Each arrow denotes one fully connected layer with a bias.}
    \label{tab:supp_policy_layers}
    \begin{tabularx}{\columnwidth}{@{}lX@{}}
        \toprule
        Component & Dimensions and output \\
        \midrule
        State encoder & $2\to256\to64$; two layers \\
        Goal encoder & $2\to256\to64$; two layers \\
        Fusion MLP & $192\to256\to256\to256$; three layers \\
        Mean head & $256\to100$;
        entries of $\mathbf{u}_{\mathrm{mean}}^{(0)}$ \\
        Variance head & $256\to100$;
        entries of $\boldsymbol{\rho}^{(0)}$ \\
        Goal-weight head & $256\to1$; $w_h$ \\
        Neural-weight head & $256\to1$; $w_n$ \\
        Temperature head & $256\to1$; $\eta$ \\
        \bottomrule
    \end{tabularx}
\end{table}

\noindent\textbf{Activations and output constraints.}
The state, goal, and fusion MLPs apply SiLU after every layer except the
last, with no additional normalization or dropout. The mean head is
linear; its outputs are clipped by the planner to $[0,1]$\,m/s for linear
velocity and $[-1,1]$\,rad/s for angular velocity. The variance head
predicts physical squared-control units and is clipped elementwise to
$[10^{-4},5]$. The three scalar heads use softplus. Since $n_h=1$,
$\boldsymbol{w}_h$ reduces to $w_h$. The network has $203$ scalar outputs
in total: $100$ means, $100$ variances, and three positive scalars.

\noindent\textbf{Head initialization.}
The variance head is initialized with zero weights and biases of $5$ for
both control channels. The goal-weight, neural-weight, and temperature
heads also have zero weights, with inverse-softplus biases producing
initial outputs $w_h=0.01$, $w_n=1000$, and $\eta=0.9$, respectively.
The mean head retains the default random linear-layer initialization.
The policy's learned position embeddings are initialized to zero.

\noindent\textbf{Parameter counts and training status.}
The collision model contains $245{,}531$ learned parameters, and the policy
contains $381{,}515$, including their respective point-cloud encoders,
temporal blocks, position embeddings, and output heads. These counts
exclude the frozen depth estimator. During policy training only
$\boldsymbol{\theta}$ is updated; the collision model is frozen and
evaluated with dropout disabled. At deployment, all parameters are fixed
and dropout is disabled in both networks.

\subsection{Training Configuration}
\label{sec:training_configuration}

Both networks are trained on an RTX~5090 with a batch size of $256$
using AdamW. We use a cosine learning-rate schedule from $10^{-4}$ to
$10^{-6}$, weight decay $6\times10^{-5}$, and global gradient-norm
clipping at $1$. Training the collision model with
$\mathcal L_{\mathrm{CM}}$ for $100$ epochs takes approximately
$45$ minutes. The collision model is then frozen. Training the policy
network through the differentiable soft-CEM layer with the combined
imitation and task-risk loss $\mathcal L$ for $100$ epochs takes
approximately one hour.

\subsection{Computation Time}
\label{sec:computation_time}

On an RTX~3080 laptop, Alg.~\ref{algo_1} alone takes $0.0035$\,s with
five soft-CEM refinements, 20 stochastic candidates, and one current-mean
anchor. Depth estimation runs in parallel and takes $0.02$\,s.
For comparison, MonoMPC's planning time is $0.016$\,s.

\section{Additional Navigation Results}
\label{sec:qualitative_comparisons}

The following subsections present qualitative results for the main-paper
experiments and an additional quantitative comparison with CARE.

\subsection{Qualitative Trajectories with MonoMPC}
\label{sec:supp_monompc}

\begin{figure*}[!t]
    \centering
    \begin{minipage}[t]{0.32\textwidth}
        \centering
        \includegraphics[width=\linewidth]{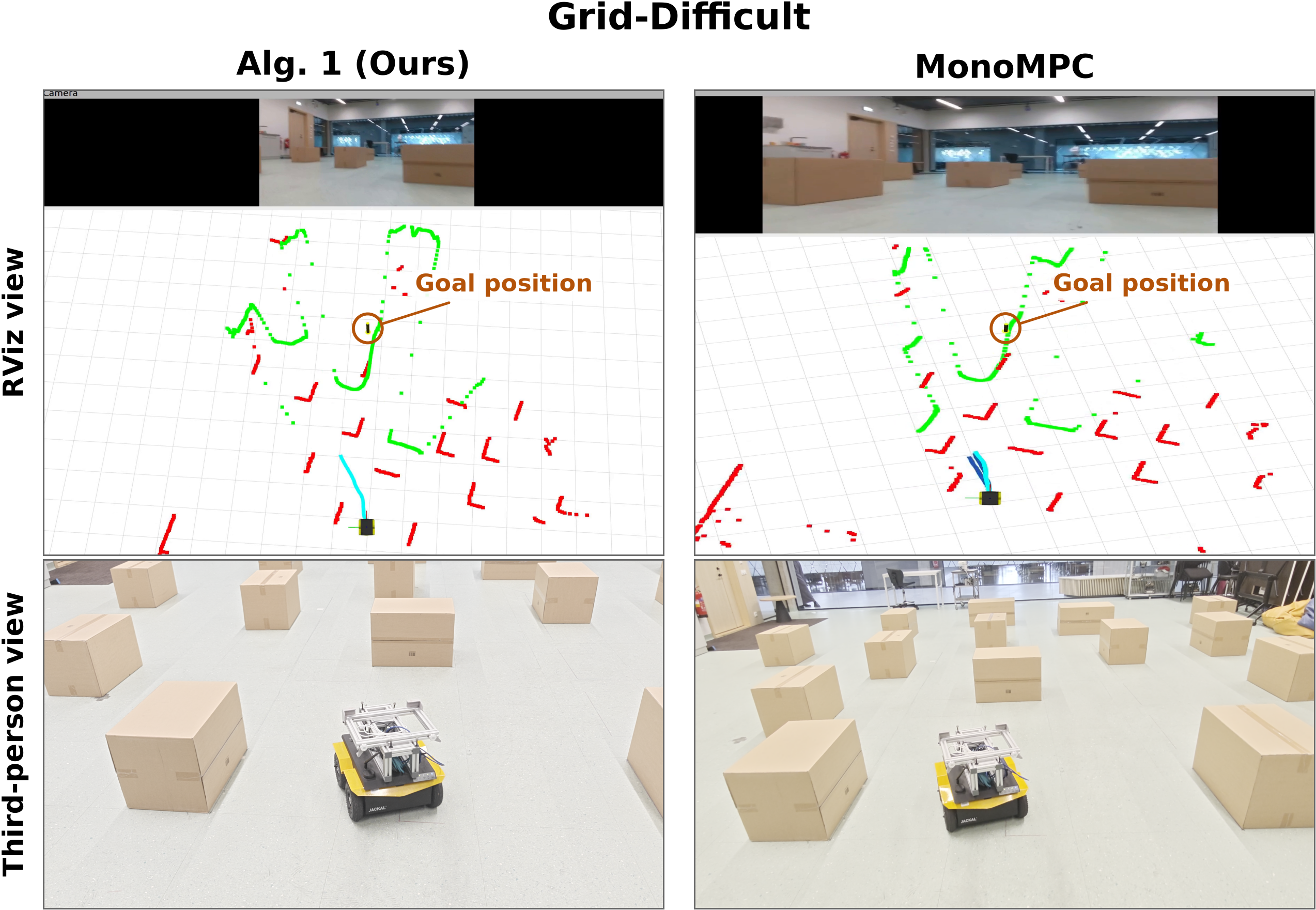}
        \par\smallskip
        (a)
    \end{minipage}\hfill
    \begin{minipage}[t]{0.32\textwidth}
        \centering
        \includegraphics[width=\linewidth]{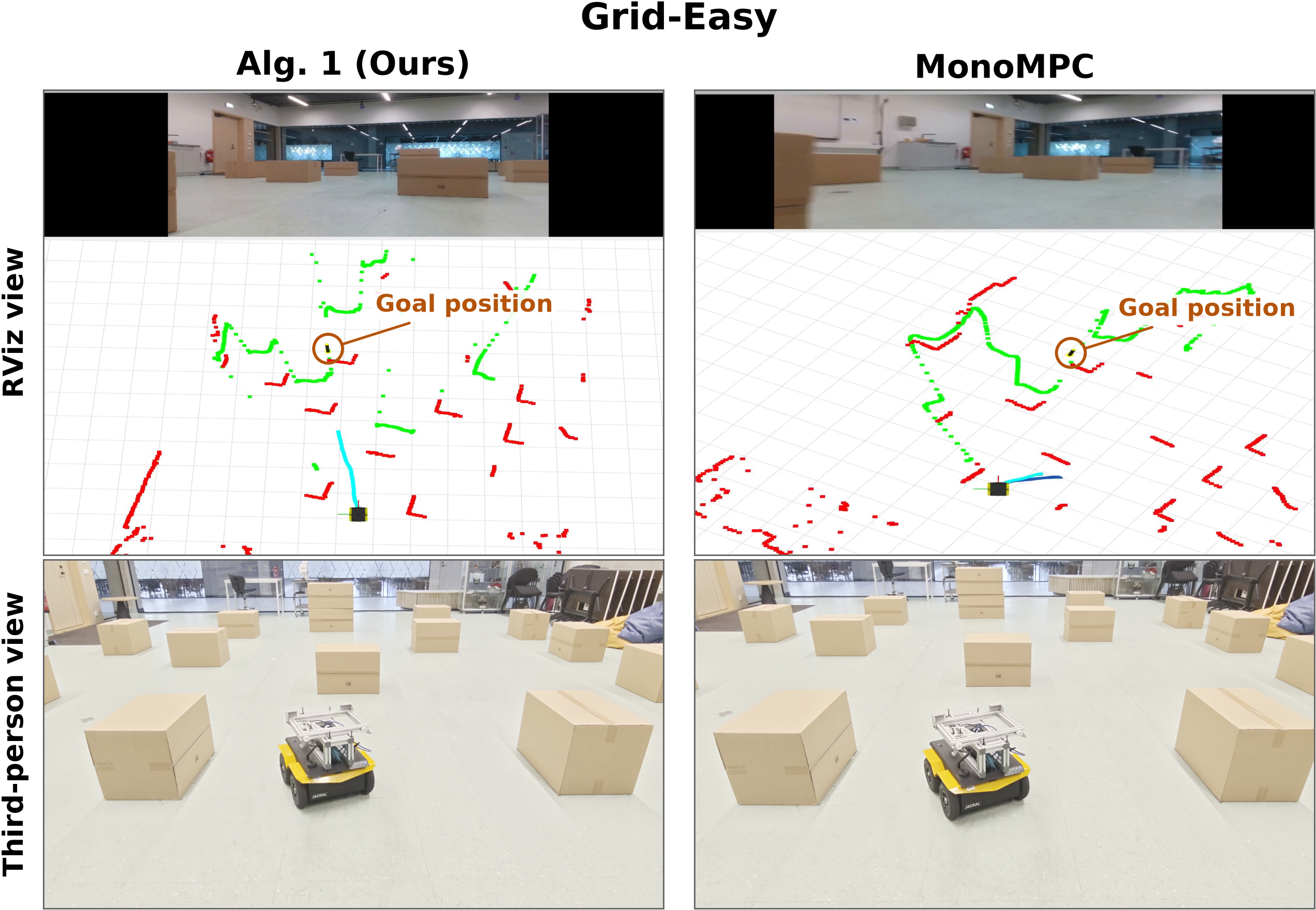}
        \par\smallskip
        (b)
    \end{minipage}\hfill
    \begin{minipage}[t]{0.32\textwidth}
        \centering
        \includegraphics[width=\linewidth]{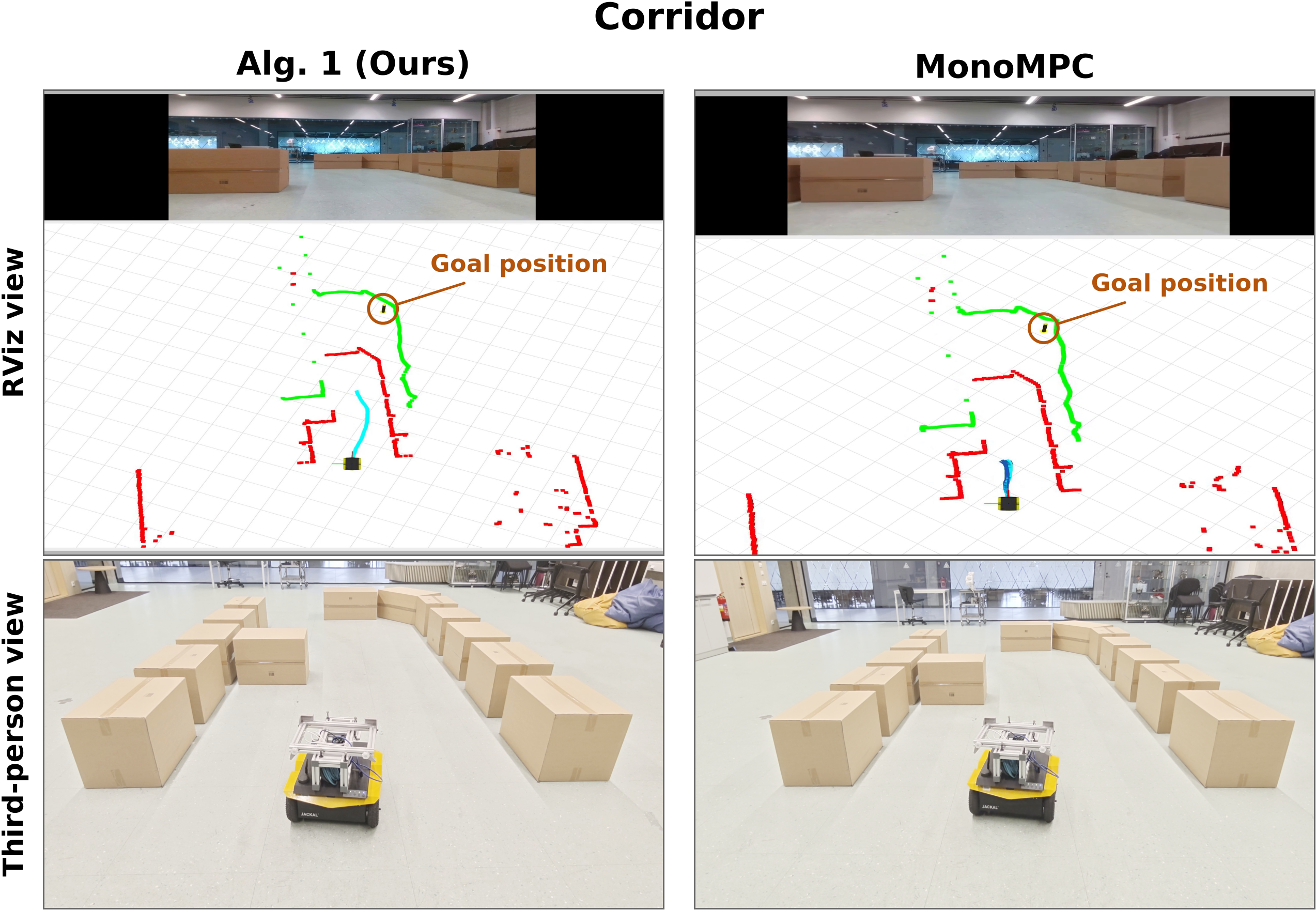}
        \par\smallskip
        (c)
    \end{minipage}
    \caption{Qualitative trajectories in (a) Grid-Difficult, (b) Grid-Easy and (c) Corridor.
    Within each panel, the left and right columns show Alg.~\ref{algo_1}
    and MonoMPC, respectively. RViz views with the goal highlighted appear
    above the corresponding third-person views.}
    \label{fig:supp_monompc_grids}
    \Description{Three panels show Grid-Difficult, Grid-Easy, and Corridor.
    Each panel contains Alg. 1 on the left and MonoMPC on the right,
    with RViz views and marked goals above third-person views.}
    \par\medskip
    \begin{minipage}[t]{0.48\textwidth}
        \centering
        \includegraphics[width=\linewidth]{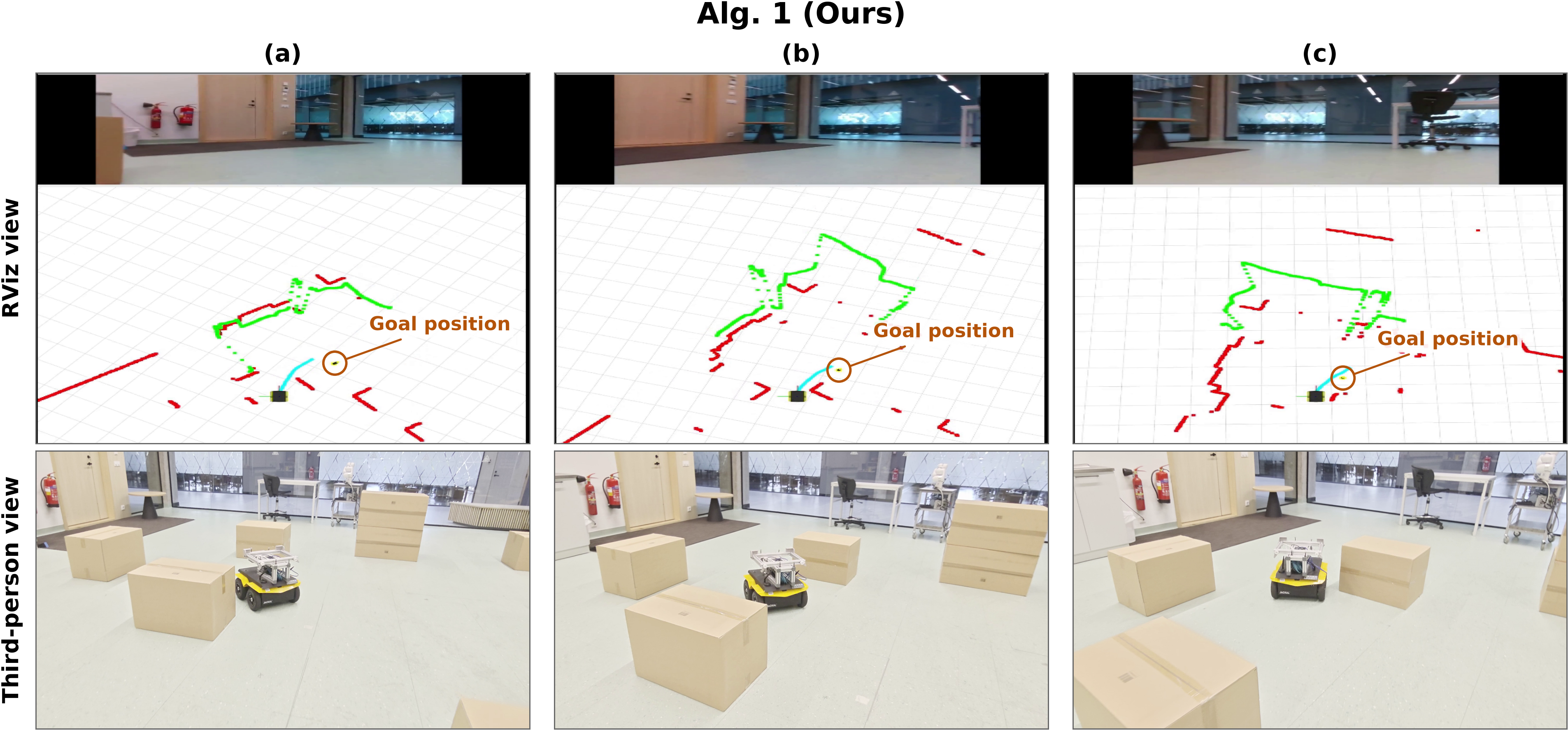}
    \end{minipage}\hfill
    \begin{minipage}[t]{0.48\textwidth}
        \centering
        \includegraphics[width=\linewidth]{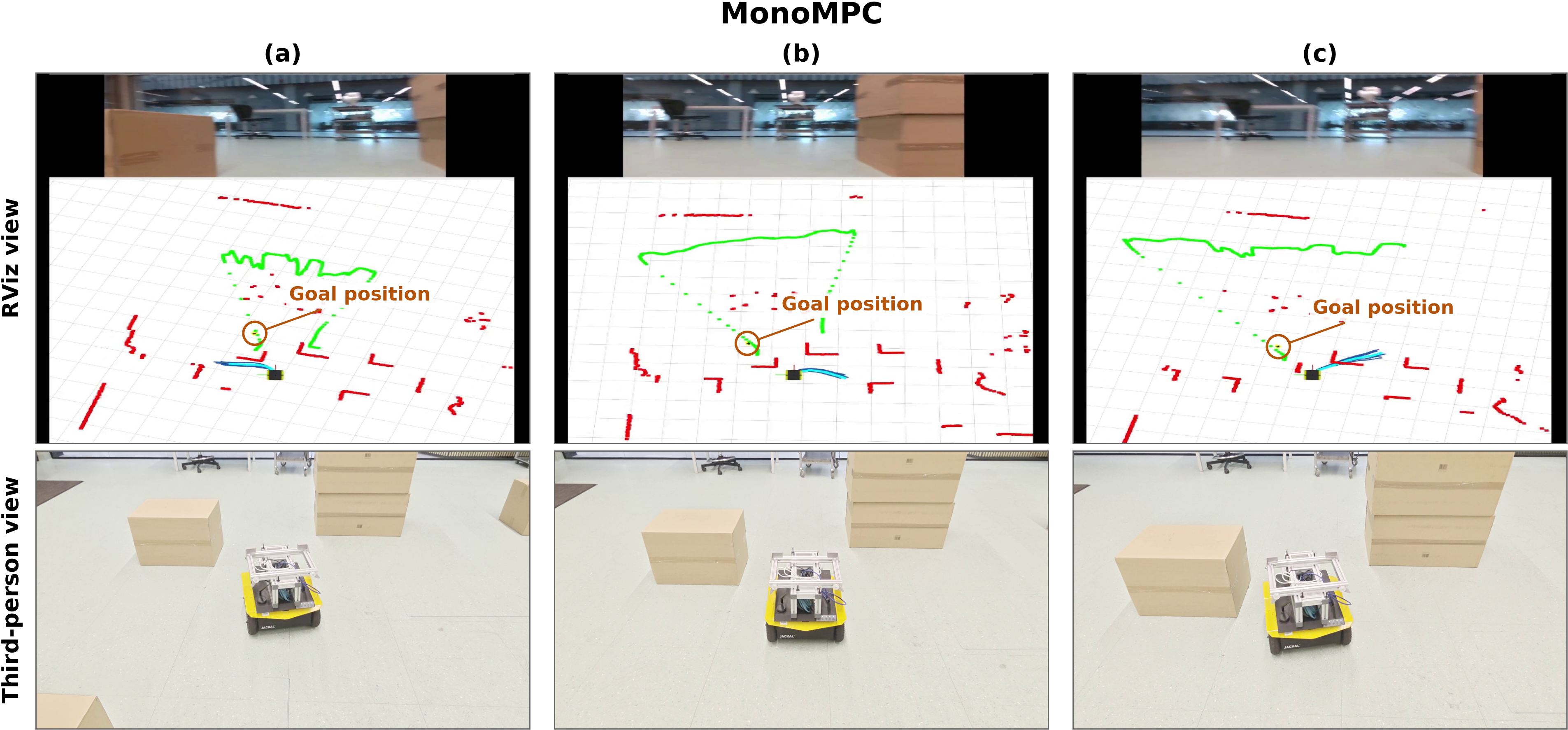}
    \end{minipage}
    \caption{Planned trajectories at selected replanning instants for
    Alg.~\ref{algo_1} (left) and MonoMPC (right).
    MonoMPC's trajectories fluctuate markedly between replanning steps,
    while our method's optimized trajectory remains stable, progresses
    toward the goal, and avoids obstacles. These snapshots are consistent with
    the lower excess heading variation and angular-velocity command
    variation reported in main-paper Fig.~\ref{fig:heading_omega}.
    Within each method, RViz views with the goal highlighted appear above
    third-person views; columns (a)--(c) are ordered chronologically.}
    \label{fig:supp_oscillations}
    \Description{Two methods are arranged side by side, each with a
    two-by-three layout showing planned trajectories at selected replanning
    instants. Alg. 1 on the left maintains a stable trajectory toward the
    goal around obstacles; MonoMPC on the right has trajectories that
    fluctuate between replanning steps. Each method has RViz views with
    marked goals above third-person views, ordered chronologically from
    left to right.}
\end{figure*}

We show qualitative plots of the trajectories generated by
Alg.~\ref{algo_1} and MonoMPC in Grid-Difficult and Grid-Easy
(Fig.~\ref{fig:supp_monompc_grids}(a) and (b), respectively) and in Corridor
(Fig.~\ref{fig:supp_monompc_grids}(c)). These plots supplement the results in main-paper
Section~\ref{sec:monompc}. The full navigation trials can be found in
the accompanying video.

\subsubsection{Oscillatory Motion}
\label{sec:supp_oscillations}

Figure~\ref{fig:supp_oscillations} shows how the planned trajectories
evolve over selected replanning instants in one of the experiments. In these
snapshots, MonoMPC's trajectories fluctuate substantially between
replanning steps, with repeated changes in the planned direction.
Alg.~\ref{algo_1}, in contrast, maintains a stable optimized
trajectory directed toward the goal while avoiding obstacles. This
behavior is consistent with the lower excess heading variation and
smaller changes in angular-velocity commands reported for our method in
main-paper Fig.~\ref{fig:heading_omega}.

The observed difference is also consistent with the use of observation
history in our collision model. MonoMPC evaluates collision risk from
the latest image, so a nearby obstacle outside the current camera view
may be omitted when a trajectory is planned. As the robot turns, the
obstacle can reappear and prompt another change in direction. Our
collision model incorporates earlier observations of such obstacles,
which can reduce repeated steering corrections and changes in the
optimized trajectory between successive MPC steps.

\subsection{Qualitative Trajectories with FlowNav}
\label{sec:supp_flownav}

\begin{figure}[ht]
    \centering
    \includegraphics[width=\columnwidth,height=0.285\textheight,keepaspectratio]{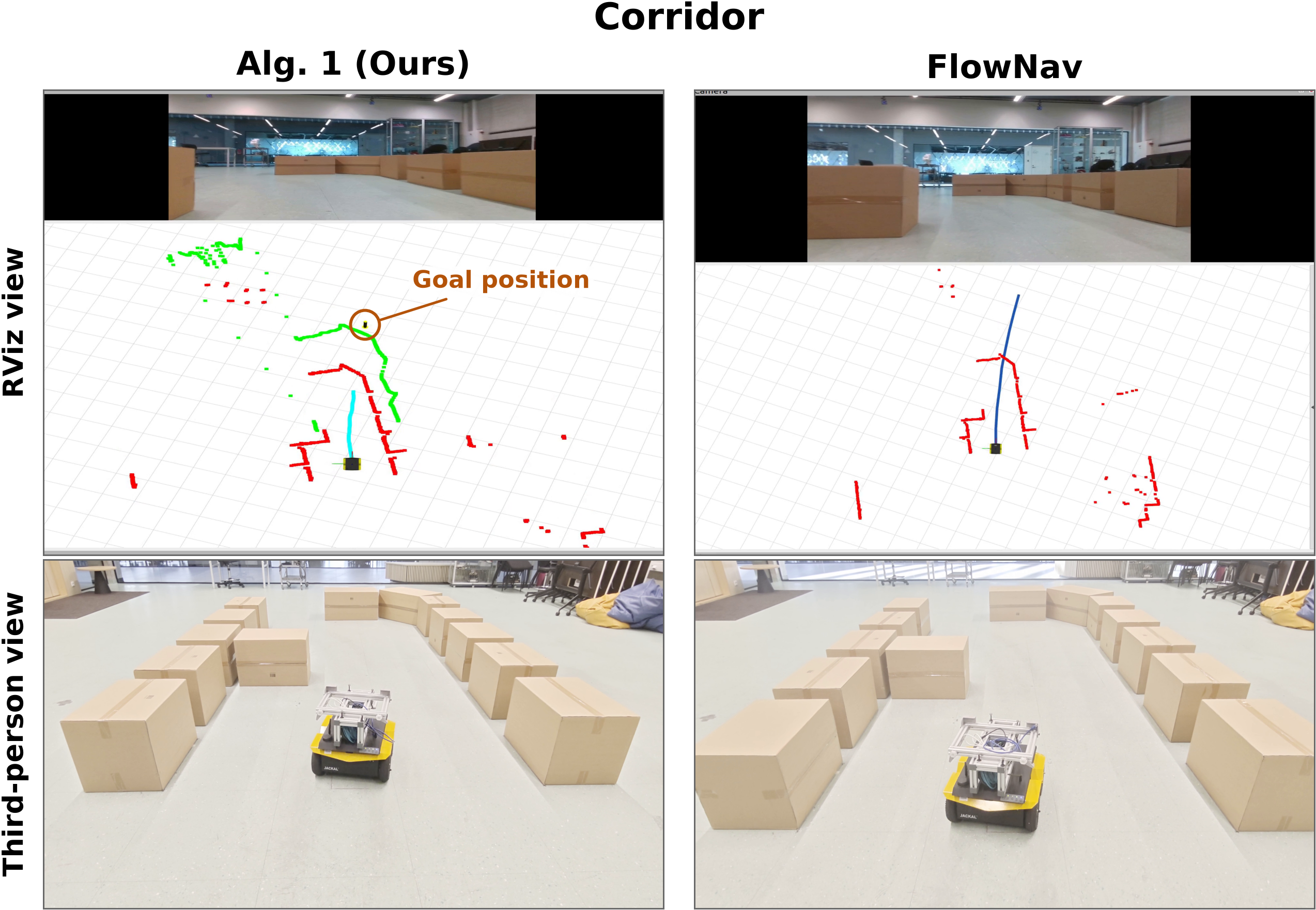}
    \caption{Alg.~\ref{algo_1} (left) and FlowNav (right) in the Corridor
    environment. The top row shows the RViz view, and the bottom row shows
    the third-person view.}
    \label{fig:supp_flownav}
    \Description{A two-by-two layout compares Alg. 1 and FlowNav in the
    Corridor environment, with RViz snapshots above third-person snapshots.
    The goal is marked in the Alg. 1 RViz view.}
\end{figure}

We show qualitative plots of the trajectories generated by
Alg.~\ref{algo_1} and FlowNav in the Corridor environment in
Fig.~\ref{fig:supp_flownav}. These plots accompany main-paper
Section~\ref{sec:flownav} and Fig.~\ref{fig:flownav}. The full navigation trials
can be found in the accompanying video.

\subsection{Comparison with CARE}
\label{sec:supp_care}

CARE~\cite{kim2025care} augments a pretrained visual-navigation policy
with depth-based repulsive corrections, without retraining the policy
or requiring additional range sensors. In our experiments, CARE is
applied to NoMaD~\cite{10610665}. It computes repulsive force vectors
from the estimated monocular depth and uses them to locally adjust
NoMaD's predicted trajectory away from nearby obstacles. This local,
hand-designed correction differs from the trajectory-level collision-risk
objective optimized by Alg.~\ref{algo_1}.

Table~\ref{tab:supp_care} reports the Corridor results. Alg.~\ref{algo_1}
achieves a collision rate of $4\%$, compared with $90\%$ for CARE,
while attaining higher average and mean per-run peak linear speeds
($0.80$ and $1.00$\,m/s versus $0.43$ and $0.57$\,m/s).
These results show that CARE's local safety corrections do not prevent
frequent collisions in the evaluated Corridor trials.

\begin{table}[ht]
    \centering
    \caption{Comparison with CARE in the Corridor environment.}
    \label{tab:supp_care}
    \begin{tabular}{@{}lccc@{}}
        \toprule
        Method & Collision & \multicolumn{2}{c}{Linear speed (m/s)} \\
        \cmidrule(lr){3-4}
        & rate (\%) & Average & Mean peak \\
        \midrule
        Alg.~\ref{algo_1} (Ours)
        & \textbf{4} & \textbf{0.80} & \textbf{1.00} \\
        CARE~\cite{kim2025care} & 90 & 0.43 & 0.57 \\
        \bottomrule
    \end{tabular}
\end{table}

\begin{figure}[ht]
    \centering
    \includegraphics[width=\columnwidth,height=0.285\textheight,keepaspectratio]{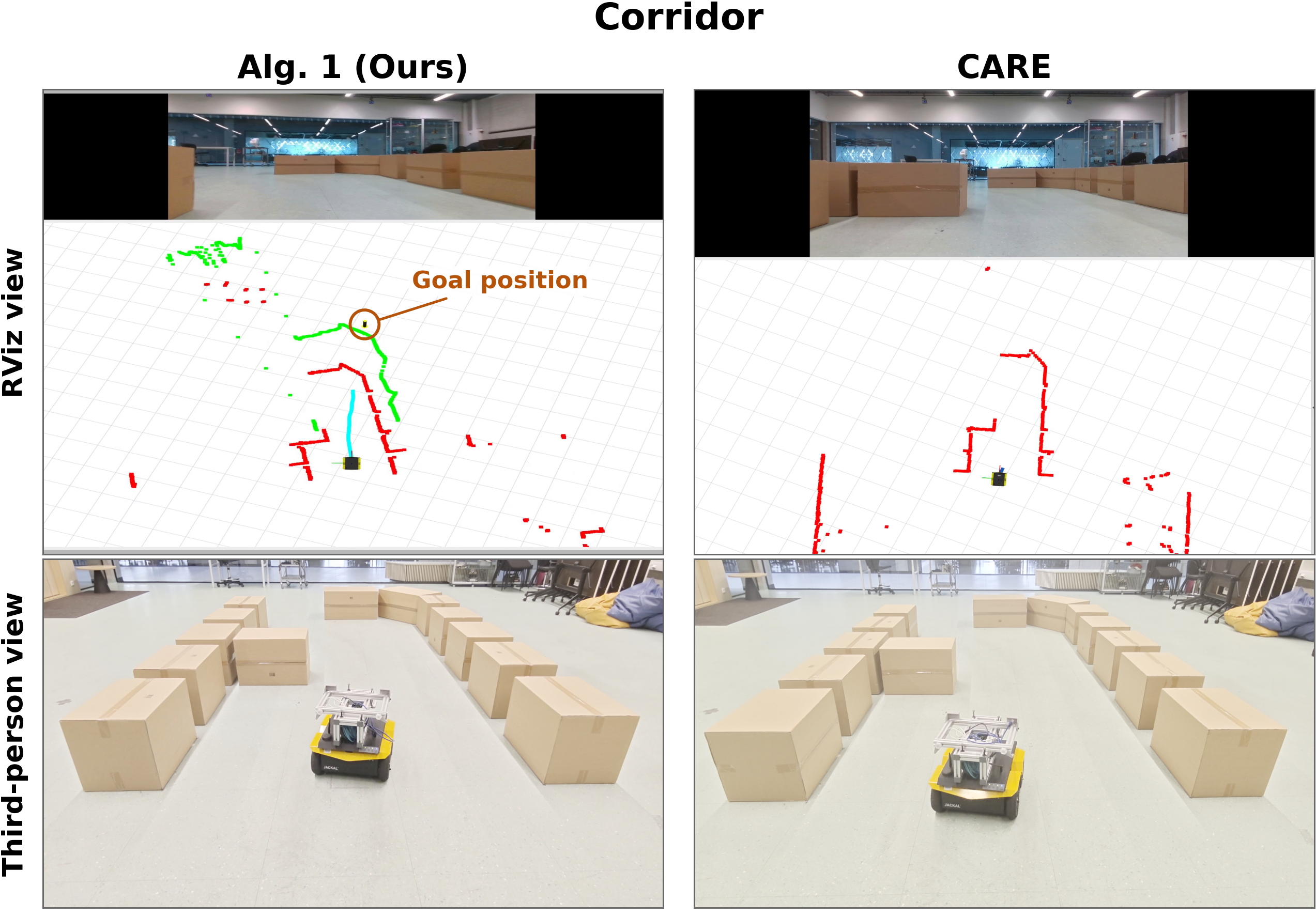}
    \caption{Alg.~\ref{algo_1} (left) and CARE applied to NoMaD (right)
    in the Corridor environment. RViz views appear above the corresponding
    third-person views. }
    \label{fig:supp_care}
    \Description{A two-by-two layout shows Alg. 1 on the left and CARE
    applied to NoMaD on the right in the Corridor environment. RViz
    snapshots appear above third-person snapshots. The Alg. 1 goal is
    marked; the CARE third-person snapshot shows the robot at an obstacle
    during a collision trial.}
\end{figure}

We show qualitative plots of the trajectories generated by
Alg.~\ref{algo_1} and CARE in Fig.~\ref{fig:supp_care}, with CARE
snapshots taken from a collision trial. The full navigation trials can
be found in the accompanying video.

\subsection{Fixed-Initialization Baselines}
\label{sec:supp_fixed_init}

\begin{figure*}[!t]
    \centering
    \includegraphics[width=\textwidth]{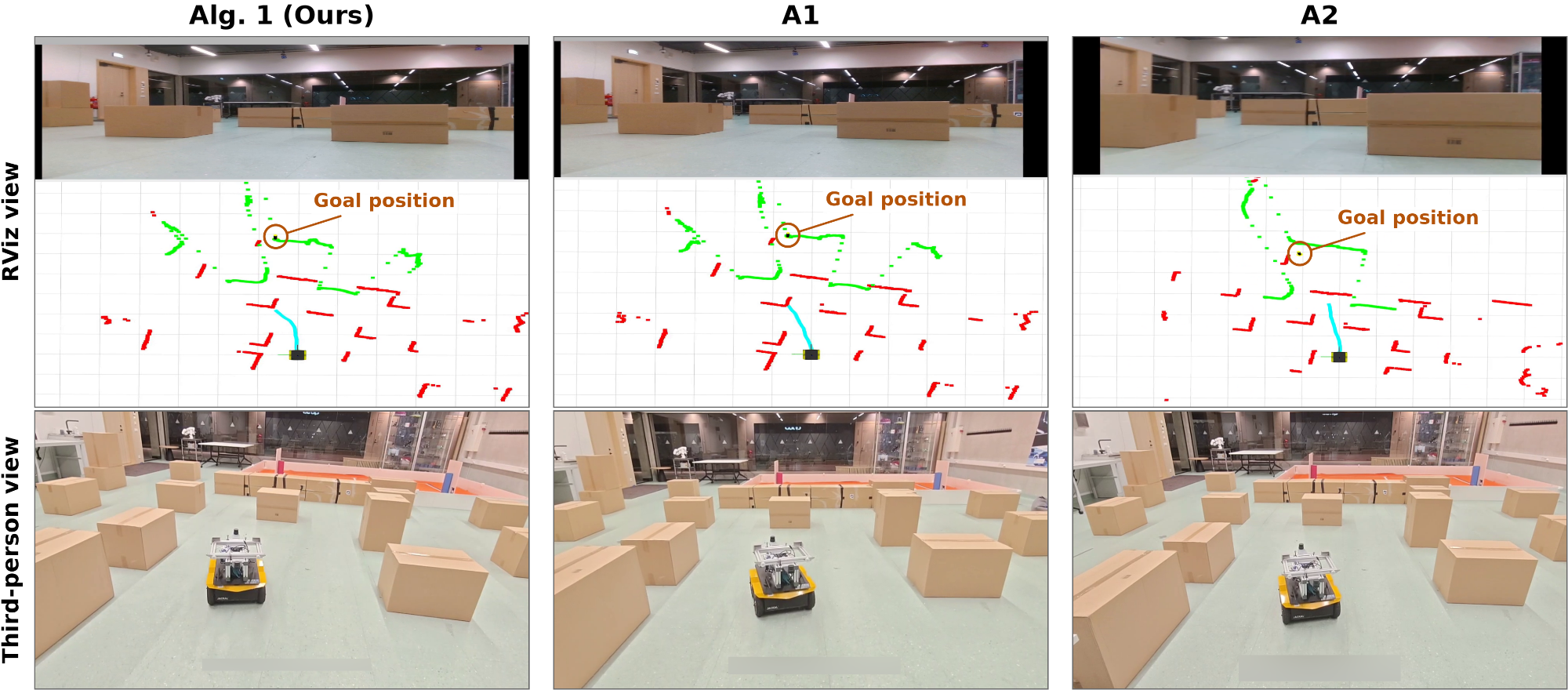}
    \caption{Alg.~\ref{algo_1} (left), A1 (center), and A2 (right).
    A1 and A2 use fixed initialization with learned and perturbed weights
    and temperature, respectively. RViz views are shown above the corresponding
    third-person views.}
    \label{fig:supp_standalone}
    \Description{A two-by-three layout compares Alg. 1 with the two
    fixed-initialization baselines. Columns show the full method, A1 with learned
    weights and temperature, and A2 with perturbed weights and temperature.
    RViz snapshots with marked goals appear above third-person snapshots.}
\end{figure*}

We show qualitative plots of the trajectories generated by
Alg.~\ref{algo_1} and the two fixed-initialization baselines, A1 and A2, in
Fig.~\ref{fig:supp_standalone}. A1 and A2 use learned and perturbed
weights and temperature, respectively, as described in main-paper Section~\ref{sec:ablation} and
Fig.~\ref{fig:ablation}. The full navigation trials can be found in
the accompanying video.

\begin{figure}[!ht]
    \centering
    \includegraphics[width=\columnwidth,height=0.39\textheight,keepaspectratio]{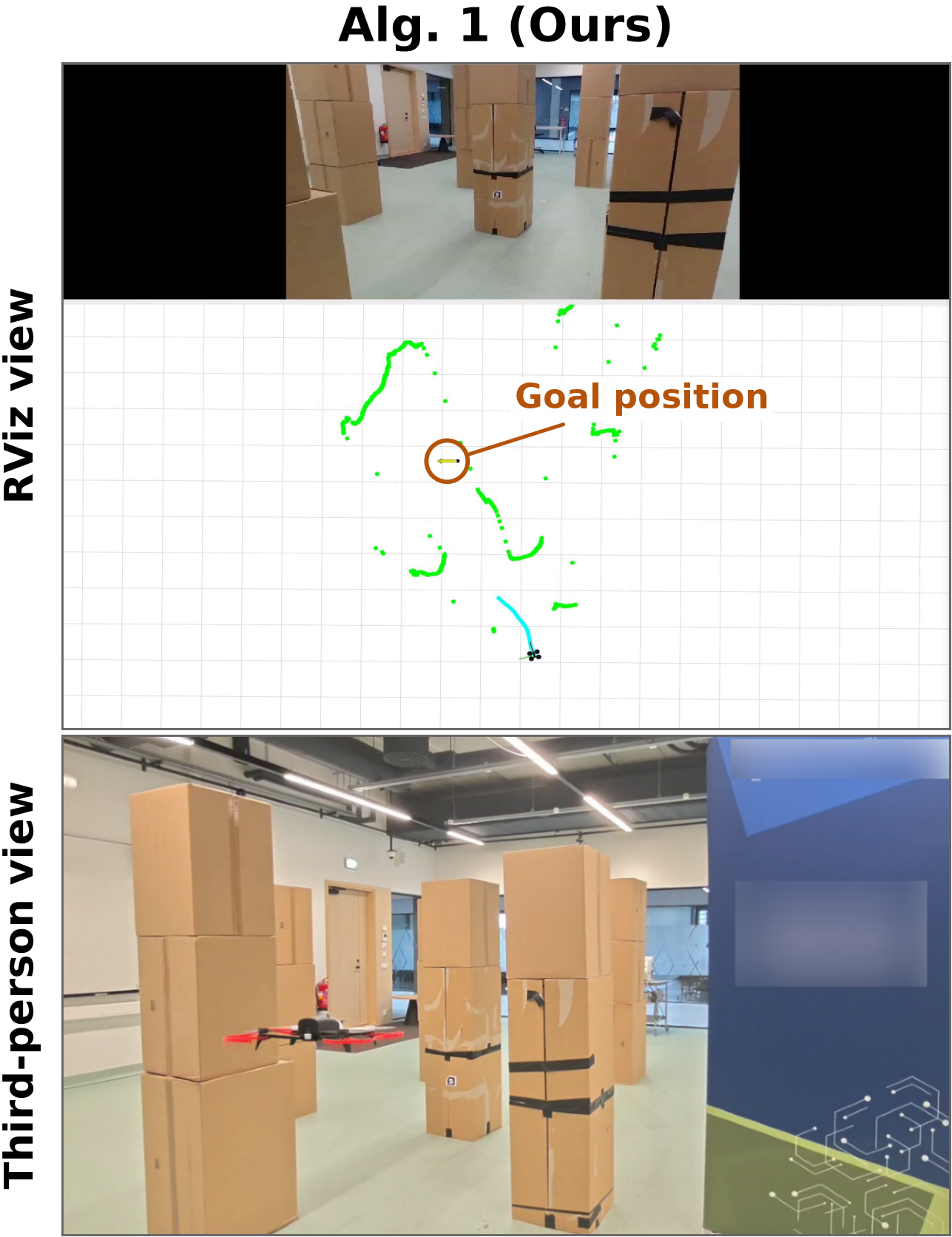}
    \caption{The complete navigation pipeline, including Alg.~\ref{algo_1},
    deployed on the Parrot Bebop without modification or retraining. The RViz view (top)
    marks the goal position, and the third-person view (bottom) shows the drone.}
    \label{fig:supp_bebop}
    \Description{Two vertically stacked snapshots show Alg. 1 on the Parrot
    Bebop. The top RViz view marks the goal, and the bottom third-person view
    shows the drone among obstacles.}
\end{figure}

\subsection{Deployment on the Parrot Bebop}
\label{sec:supp_bebop}

We show qualitative plots of the trajectories generated by
Alg.~\ref{algo_1} during deployment on the Parrot Bebop in
Fig.~\ref{fig:supp_bebop}. The entire navigation pipeline transfers without
modification or retraining, as described in main-paper Section~\ref{sec:zero_shot}. The full navigation trials can be
found in the accompanying video.

\section{Evaluation Metrics}
\label{sec:evaluation_metrics}

The metrics below are computed from the corresponding method- and
environment-specific deployment logs. At each recorded step $t$, the
logs contain linear velocity $v_t^{\mathrm{lin}}$, angular velocity
$v_t^{\mathrm{ang}}$, and heading $\vartheta_t$.

\subsection{Collision Rate and Speed}

Collision rate is the percentage of navigation runs in which a collision
occurs. For each run, average linear speed is the arithmetic mean of
$v_t^{\mathrm{lin}}$ over all recorded steps, and peak linear speed is
the maximum recorded linear velocity. The reported average speed and
mean per-run peak speed are the arithmetic means of these respective
per-run statistics over all available runs. Each run has equal weight,
regardless of its duration. Both linear-speed metrics are reported in
m/s. For angular speed, the same averaging and peak operations are
applied to $|v_t^{\mathrm{ang}}|$, with results reported in rad/s.
Speed statistics are interpreted together with collision rate to assess
whether faster motion compromises safety.

\subsection{Steering Oscillation}

For the MonoMPC comparison, we additionally quantify back-and-forth
heading fluctuations and changes in angular-velocity commands. Both
methods use the same logging interval, $\Delta t_{\mathrm{log}}=0.1$\,s.
For a run with $T\geq2$ recorded steps, let $\widetilde\vartheta_t$ denote
the unwrapped heading at step $t$. Excess heading variation is
\begin{equation}
    \overline{\Delta\vartheta}
    =\frac{
        \displaystyle\sum_{t=0}^{T-2}
        |\widetilde\vartheta_{t+1}-\widetilde\vartheta_t|
        -|\widetilde\vartheta_{T-1}-\widetilde\vartheta_0|
    }{T-1}.
    \label{steering_metrics}
\end{equation}
Subtracting the absolute net heading change removes the turn required
by the route; the remainder quantifies back-and-forth fluctuations.
Dividing by $T-1$ accounts for different run lengths, giving units of
rad/step. Mean absolute angular-velocity command variation is
\begin{equation}
    \overline{|\Delta v^{\mathrm{ang}}|}
    =\frac{1}{T-1}\sum_{t=0}^{T-2}
    |v_{t+1}^{\mathrm{ang}}-v_t^{\mathrm{ang}}|.
    \label{angular_command_variation}
\end{equation}
This statistic measures the mean absolute command change per logged
transition and is reported in rad/s. It is not divided by the logging
interval and therefore is not angular acceleration. Both metrics are
computed separately for every run, and their per-run distributions are
shown for Corridor, Grid-Easy, and Grid-Difficult in main-paper
Fig.~\ref{fig:heading_omega}. Pooled averages give each run equal
weight. Lower values indicate less heading oscillation and less abrupt
steering input, respectively.

\end{document}